\documentclass[12pt]{article}

\usepackage[left=3.4cm,right=3.4cm,bottom=3.5cm,top=3.4cm]{geometry}
\usepackage{graphicx}
\usepackage{amsmath,amsfonts}
\usepackage[utf8]{inputenc} 
\usepackage[T1]{fontenc}    
\usepackage{hyperref}       
\usepackage{url}            
\usepackage{booktabs}       
\usepackage{nicefrac}       
\usepackage{microtype}      
\usepackage{xcolor}         
\usepackage{bm}
\usepackage[numbers]{natbib}
\usepackage{comment}
\newcommand{\red}[1]{\textcolor{black}{#1}}

\renewcommand{\thefigure}{\thesection.\arabic{figure}}
\renewcommand{\thetable}{\thesection.\arabic{table}}
\usepackage{chngcntr}
\counterwithin{figure}{section}
\counterwithin{table}{section}

\title{Multi-modal cascade feature transfer for polymer  property prediction}

\author{Kiichi Obuchi\textsuperscript{1,2}, Yuta Yahagi\textsuperscript{1,2}, Kiyohiko Toyama\textsuperscript{2},\\ Shukichi Tanaka\textsuperscript{2}, and Kota Matsui\textsuperscript{3}
}
\date{}

\begin{document}
\maketitle

\noindent
\textsuperscript{1}NEC Corporation, Minato-ku, Tokyo, Japan, 108-8001 \\
\textsuperscript{2}National Institute of Advanced Industrial Science and Technology, Tsukuba, Japan, 305-8568 \\
\textsuperscript{3}Kyoto University, Kyoto, Japan, 606-8501 \\
\\
\texttt{Email:kiichi-obuchi@nec.com}



\begin{abstract}
\red{In this paper, we propose a novel transfer learning approach called multi-modal cascade model with feature transfer for polymer property prediction. 
  Polymers are characterised by a composite of data in several different formats, including molecular descriptors and additive information as well as chemical structures.
  However, in conventional approaches, prediction models were often constructed using each type of data separately. 
  Our model enables more accurate prediction of physical properties for polymers by combining features extracted from the chemical structure by graph convolutional neural networks (GCN) with features such as molecular descriptors and additive information.
  The predictive performance of the proposed method is empirically evaluated using several polymer datasets. We report that the proposed method shows high predictive performance compared to the baseline conventional approach using a single feature.}
\end{abstract}
 \normalsize

\section{Introduction}

Polymers are attractive material targets with a wide range of applications. Databases for polymers are being developed mainly in the scientific community~\citep{otsuka2011polyinfo,kim2018polymer,hayashi2022radonpy}, and some predictive model developments for polymers have been reported~\citep{martin2023emerging,kuenneth2022bioplastic,wu2019machine,tao2021benchmarking}.
Such models can be used, for example, to consider the application of adaptive experimental design~\citep{8957442, Garud2017-zd}, an effective data-driven approach. Adaptive experimental design is a method of repeated data collection and model updating using machine learning models trained on the data, which allows target characteristics to be reached quickly.

However, the use of machine learning models in the polymer domain faces the following two challenges: (1) The cost of data collection in chemical fields is expensive and often sufficient data is not accumulated, especially in the field of advanced materials. In such cases, it is difficult to pre-train high-performing machine learning models (the so-called cold start problem). 
(2) As polymers contain not only molecular structure but also other information (molecular weight, additional information, etc.), a multi-modal model is needed that can adequately represent molecular structure and simultaneously process non-structural features such as numbers and vectors.

Transfer learning~\citep{zhuang2020comprehensive,Yang_Zhang_Dai_Pan_2020} is a promising machine learning approach to address the aforementioned issues (1) and (2). 
Transfer learning refers to a methodology that leverages abundant data or other forms of knowledge from a different domain to enable efficient machine learning in a domain with limited data. 
In general, the domain where the target task is defined (referred to as the target domain) differs from the domain from which knowledge is borrowed (referred to as the source domain). 
Consequently, transfer learning naturally necessitates the consideration of multimodality~\cite{zhuang2020comprehensive, zhen2020deep}. 
Transfer learning can be broadly categorized into three approaches based on the perspective of "what to transfer": (i) {\it instance transfer}~\cite{sugiyama2012density}, which involves transferring the data itself from the source domain, (ii) {\it feature transfer}~\citep{dai2008translated, ganin2016domain, he2023cross}, which utilizes features obtained through representation learning, and (iii) {\it parameter transfer}, which transfers pre-trained models~\cite{ding2023parameter}. 
In this study, we employ the feature transfer approach to address transfer learning between data with different formats, specifically graph and tabular data. 
This is because instance transfer is unsuitable in this setting, as data from the source domain with a different format cannot be directly combined with and used alongside data from the target domain. 
Similarly, pre-trained models are generally designed for transfer between data of the same format, making parameter transfer an inappropriate approach for this study.

This paper proposes a multi-modal cascade model with feature transfer as a method for developing predictive models for polymers. \red{Our model is a machine learning architecture developed with consideration of polymer properties. Specifically, it is assumed by polymer scientists that polymer properties can be described by two distinct components: one represented by molecular structures of monomers and the other by additional vector representations. To integrate these components into a unified predictive framework, we employed a cascade modeling approach, enabling a more structured and interpretable representation of polymer properties. From the perspective of multi-modality,} our approach combines different features, which originally input from different formats, in a cascaded model structure to achieve a more effective and flexible prediction. Furthermore, if pre-trained models can be used in the feature extractor of the cascade model, it is possible to improve the accuracy of the model in small data domains. This is expected to enable the use of big data from the public and private sectors, including past experiments, for effective modeling of private small-scale data in clandestine projects. \red{Regarding industrial application, in material design for the plastics and rubber industries, two major aspects are polymer design and formulation design, which involves adding additives. Our proposed method can be applied to both. For example, it can be used in the polymer design for functional polymer as well as in cases where other additives are incorporated into polymers for additional function. Since these scenarios often suffer from limited data, utilizing pre-trained models in a cascade approach can help build high-accuracy models, enabling the rapid development of desired materials with fewer experiments. This contributes to accelerating materials development.}

\section{Related Work}
Predictive modelling of polymer properties from molecular structures is mainly done by machine learning models such as decision trees and all-connected Neural Networks, after converting them into molecular descriptors, language models using strings such as SMILES(simplified molecular input line entry system), and GNNs (Graph Neural Networks), which regard molecular structures as molecular graphs.
Tao et al.~\citep{tao2021benchmarking} compare 79 different machine learning models for polymer Tg(glass transition temperature) and compare their characteristics. There are also reports that some attempt to overcome the challenge of small data using transfer learning. Wu et al.~\citep{wu2019machine} constructed an all-connected network and transferred it to predict thermal conductivity. Kuenneth et al.~\citep{kuenneth2022bioplastic} built a molecular graph-based multitask learning model to predict multiple physical properties. \red{Several Transformer-based models have recently published.} Kuenneth et al.~\red{\citep{kuenneth2023polybert} proposed a pipeline that includes a polymer chemical fingerprinting capability called polyBERT, and a multitask learning approach that maps the polyBERT fingerprints to host of properties.} Xu et al.~\citep{xu2023transpolymer} proposed a model that can be applied to copolymers using a language model(TransPolymer) and demonstrated it as a predictive model for electrical properties of calculated data from DFT, such as conductivity and band gap.
Zhou et al.~\citep{zhou2025polycl}utilised unlabelled data to build a learning model for the language model(PolyCL), using a self-supervised contrastive learning, and predicted physical properties in their downstream task.

The feature transfer approach employed in this study can be further divided into two implementations: (1) transferring features pre-developed through representation learning in the source domain to the target domain, and (2) performing representation learning to obtain features that can be commonly utilized in both the source and target domains (such features are referred to as domain invariant features). 
In the former approach, features $\bm{f}_s$ and $\bm{f}_t$ are first learned independently through representation learning in the source and target domains, respectively, and a prediction model is trained using the combined features $[\bm{f}_s,\bm{f}_t]$~\citep{hoffman2013efficient, kulis2011you, qi2011towards}. 
In contrast, the latter approach involves learning invariant features $\bm{f}$ from the input data $\bm{x}_s$ and $\bm{x}_t$ of both domains, and using these features to train the prediction model~\citep{daume2007frustratingly, zhuang2015supervised, ganin2016domain}. 
Domain-invariant representation learning is an attractive approach; however, creating useful invariant features requires training a deep neural network model with the assumption of abundant data in both domains. 
Since this requirement is incompatible with the setting of this study, which assumes limited data in the target domain, we adopt the former approach here. 
In the former approach, prior to the advent of deep learning, it was common to perform processing that mitigated the discrepancy between domains when combining features created in each domain. 
However, it has now become standard practice to utilize representation learning with pre-trained models~\citep{han2021pre} for this purpose. 

\red{The comparison chart with other related models is illustrated in Table~\ref{fig:comparison_chart}, which are compared regarding pretrained-model utilization, multi modality for polymer complexity, compatibility with other models and datasets.The novelty of this study lies in the cascade model. This proposed approach is expected to enable the utilization of pre-trained models while allowing for multi-modal inputs from various sources. Additionally, rather than competing with other methods, our architecture is designed to integrate and leverage existing approaches, making it highly compatible with state-of-the-art pre-trained models.}

\begin{table}[t]
 \caption{Comparison chart with other related models
 \label{fig:comparison_chart}
 \vspace{-0.5cm}
 }
 \begin{center}
 \begin{tabular}{c}
  
 \hspace{-0.5cm}
  \includegraphics[clip, width=1.0\columnwidth]{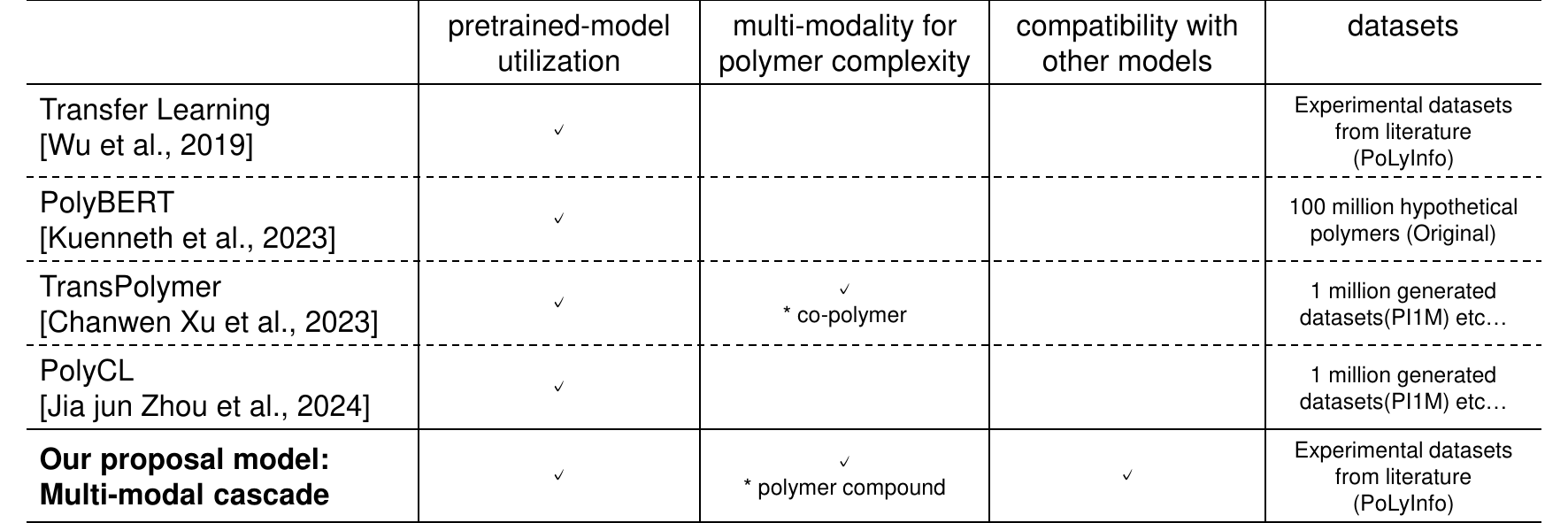}
 \end{tabular}
 \end{center}
\end{table}

\section{Methods}
\label{gen_inst}

\subsection{Data format and conventional approach}

The chemical structures treated in this study were assumed to consist of node index: atom number, atom mass, ring and aromatic/edge attribution: bond type (single, double, triple, aromatic), ring and aromatic/adjacent matrix, and was assumed to be described in SMILES format. 
In addition to the chemical structure, a feature vector encoding the molecular descriptors by  RDKit and  information of additives encoded in an one-hot vector was used. 
While previous approaches have involved building a separate prediction model for each data format (left-hand side of Figure~\ref{fig:conbentional-vs-proposal}), this study proposes a framework in which data from all formats are processed in a unified manner using a single model.

\subsection{Proposed Methods}

\begin{figure}[t]
 \begin{center}
 \begin{tabular}{c}
  \includegraphics[clip, width=1.0\columnwidth]%
  {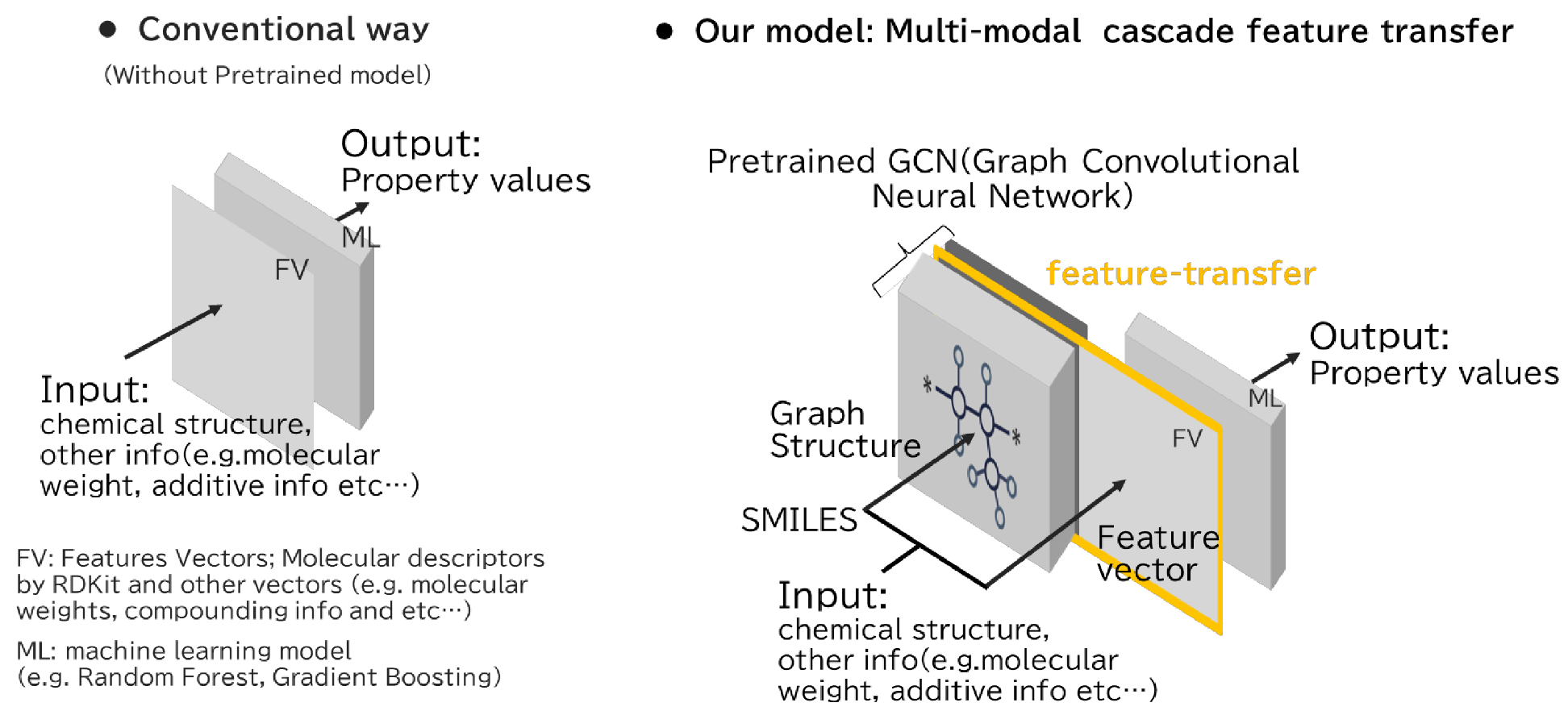}
 \end{tabular}
 \caption{Conventional way (benchmark) and Our model 
 \label{fig:conbentional-vs-proposal}
 }
 \end{center}
\end{figure}

Our model is constructed in the form of a combination of two models.
The first model is a feature extractor for chemical structures and in this study a graph convolutional neural network (GCN)~\citep{schlichtkrull2018modeling} is employed. 
The second model predicts physical properties for polymers by using features from GCNs together with features from molecular descriptors and additives.
We represent these two models as a single model by combining them by means of a cascade structure (right-hand side of Figure~\ref{fig:conbentional-vs-proposal}). 

\begin{figure}[t]
 \begin{center}
 \begin{tabular}{c}
  \includegraphics[clip, width=1.0\columnwidth]{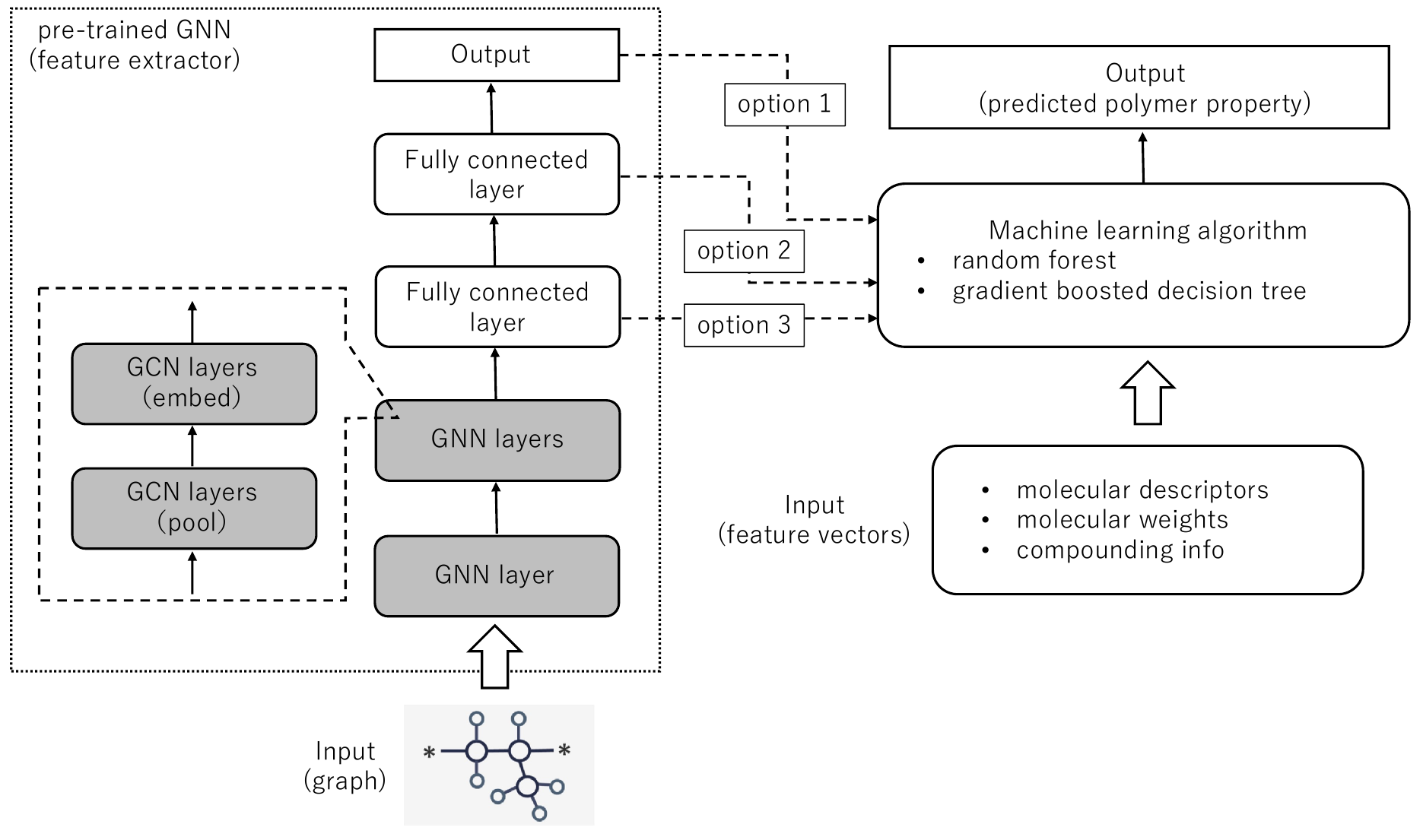}
 \end{tabular}
 \caption{Detailed structure of the proposed pipeline for a polymer property prediction model. 
 First, feature extraction is performed from molecular structure data represented as graphs using a pre-trained GNN model (left side of the figure). The extracted features, along with other data such as molecular descriptors and compounding information, are then used as inputs to a machine learning algorithm to obtain predicted values for the desired polymer properties (right side of the figure).
 \label{fig:proposed_methods}
 }
 \end{center}
\end{figure}

GCNs is a representative architecture of GNNs based on message passing. 
Specifically, they are neural network models that generate an embedding vector $\bm{z}$ for the node features $\bm{X}$ as described below. 
The input graph $G = (V, E, \bm{X})$ is defined by the vertex set $V$, the edge set $E$, and the node features $\bm{X}$. 
Then our $L$-layer GCNs transform the node features $\bm{X}_v$ of a vertex $v$ into an embedding vector $\bm{z}_v$ and output it as follows: 
\begin{align*}
\begin{aligned}
\mbox{Input layer:~~~}\bm{h}_v^{(0)} &= \bm{X}_v, \\ 
\mbox{GCN layer:~~~}\bm{h}_v^{(l)} &= \sigma \left(\sum_{u \in \mathcal{N}(v)} \frac{1}{\sqrt{\mathcal{N}(v)}\sqrt{\mathcal{N}(u)}} (\bm{W}^{(l)}\bm{h}_u^{(l-1)} + \bm{b}^{(l)})  \right), 1 \le l \le L-m \\
\mbox{Linear layer:~~~}\bm{h}_v^{(l)} &= \bm{W}^{(l)}\bm{h}_v^{(l-1)} + \bm{b}^{(l)}, L-m \le l \le L \\ 
\mbox{Output layer:~~~} \bm{z}_v &= \bm{h}_v^{(l)}
\end{aligned}
\end{align*}
where $\mathcal{N}(v) = \{u \in V \mid \{v, u\} \in E\}$ is the neighborhood vertex set of $v$, $\bm{W}^{(l)} \in \mathbb{R}^{d_l \times d_{l-1}}$ and $\bm{b}^{(l)} \in \mathbb{R}^{d_l}$ are parameter matrix and bias vector of $l$-th layer, and $\sigma$ is an activation function such as sigmoid or ReLU. 
The specific architecture is shown on the left side of Figure~\ref{fig:proposed_methods}.

Our approach aims to train a predictive model by integrating graph-structured data with tabular data, which are represented in different formats. 
To achieve this, the graph data is transformed into feature vectors using the GCNs described above. 
These feature vectors are then combined with the tabular data to serve as inputs to the prediction model. 
This process can be regarded as feature transfer from the domain of graph-structured data to the domain of tabular data.

It is worth noting the flexibility in selecting which layer's output from the GCNs to transfer as features. 
\red{
In general, it is known that in neural network models, features obtained from layers closer to the output layer tend to be more task-specific, while those from layers closer to the input layer are more general-purpose.
Although the latter may appear more suitable for transfer learning at first glance, models utilizing overly generic features often do not exhibit high performance in practice. 
Indeed, it has been reported that the performance can vary depending on the choice of the transferred layer, even for the same task~\citep{huang2022frustratingly}.
For this reason, we consider options regarding which layer of the GCNs the prediction model should be coupled with.
In this study, considering the nature of the task of predicting polymer properties, we judged that features extracted from layers closer to the output layer would function effectively.
}
Specifically, we consider three options formulated as following Eq.~\eqref{eq:option1} and Eq.~\eqref{eq:option2}. 
Option 1 connects the $L$-th (final) layer of the GCN ($L$-th layer) with the prediction model, whereas option 2 and 3 conect the $L-1$ and $L-2$ layer with the prediction model: 
\begin{align}
    \label{eq:option1}
    \mbox{option~1:~~~}
    \bm{F}^{(L)} 
    &= 
    [\bm{H}^{(L)}, \bm{X}^{\mathrm{tab}}], ~~~
    \bm{H}^{(L)} = (\bm{h}_v^{(L)})_{v \in V}, \\ 
    \label{eq:option2}
    \mbox{option~2, 3:~~~}
    \bm{F}^{(l)} 
    &= [\bm{H}^{(l)}, \bm{X}^{\mathrm{tab}}], ~~~
    \bm{H}^{(l)} = (\bm{h}_v^{(l)})_{v \in V}, ~~~ l = L-1, L-2
\end{align}
where, for option~1, $\bm{H}^{(L)}$ are extracted features from GCNs of $L$-th layer, $\bm{X}^{\mathrm{tab}}$ are feature vectors such as molecular descriptors by RDKit and other vectors (e.g. molecular weights, compounding information, etc.). 
Furthermore, $[\bm{a}, \bm{b}]$ is the symbol for the concatenation of vectors $\bm{a}$ and $\bm{b}$. 
The basic symbol definition is the same in option 2 and 3. 
Finally, we use the new feature vector $\bm{F}^{(l)}$ $l = L, L-1, L-2$ as the input for our prediction model. 
The overall structure of the proposed method's pipeline is illustrated in Figure~\ref{fig:proposed_methods}. \\
\red{In this study, The GCN model consists of three GCN Conv layers with batch normalization and ReLU activation. The DiffPool module hierarchically coarsens the graph across two levels, reducing the number of nodes to 25\% at each stage. It includes two pooling and two embedding networks. The final output is processed through two fully connected layers (Linear(64, 64) and Linear(64, 1)), making the model suitable for regression tasks.}

\red{As an additional study, we attempt to integrate our approach with a state-of-the-art model to demonstrate its compatibility. PolyCL~\citep{zhou2025polycl} is selected as a case study because it is one of the most recent Transformer-based models for polymer prediction. 
We followed the instructions provided in their GitHub repository and obtained 600 vectors from each SMILES representation. These vectors were used as input for the machine learning algorithm along with other feature vectors, instead of features derived from the GNN.}
  
\subsection{Data Preparation and Cleansing}
In this study, as a proof of concept of polymer case, we prepared datasets by following steps.

Polymer datasets were prepared by filtering in PoLyInfo~\citep{otsuka2011polyinfo} regarding neat resin and compound datasets of simple structure, which had \red{glass transition temperature:} Tg property. \red{PoLyInfo is a database that extracts property data from the literature. As of March 26, 2025, it contains property data for the following polymers.}
\textit{(Reference link}: \url{https://polymer.nims.go.jp/datapoint.html)}
{
\noindent

\noindent
    \centering
    \begin{tabular}{p{0.45\textwidth} p{0.45\textwidth}}  
        \begin{itemize}
            \item Homopolymers: 19,191
            \item Copolymers: 8,263
            \item Polymer Blends: 2,766
            \item Composites: 3,177
        \end{itemize}
        &
        \begin{itemize}
            \item Polymer sample: 173,817
            \item Data points: 546,753
            \item References: 21,684
        \end{itemize}
    \end{tabular}

\noindent
}
\red
{The glass transition temperature (Tg) is a crucial physical property in polymer design, as it significantly influences mechanical properties such as flexibility, impact resistance, and tensile strength. Furthermore, it also affects other performance aspects, including biodegradability and long-term stability. From a data science perspective, a substantial amount of data related to glass transition temperature has been made publicly available, making it an attractive target for machine learning applications.} 
Then we operated data cleansing on compounding datasets because additives information was not tidy. For example, several pieces of information were contaminated in one cell, such as additives' names, values, and units. Sometimes these ways of term separation were not uniform. Additionally, several additives made it more complicated. Under these complicated situations, rule-based cleansing did not work well. Thus, we tried using LLM (Claude 3.5 Sonnet~\citep{sonet}) in order to curate these datasets interactively. It resulted in 7503 neat resin datasets(NR datasets) and 453 compounding datasets (Comp datasets). 

\begin{figure}[t]
 \begin{center}
 \begin{tabular}{c}
  \includegraphics[clip, width=1.0\columnwidth]
  {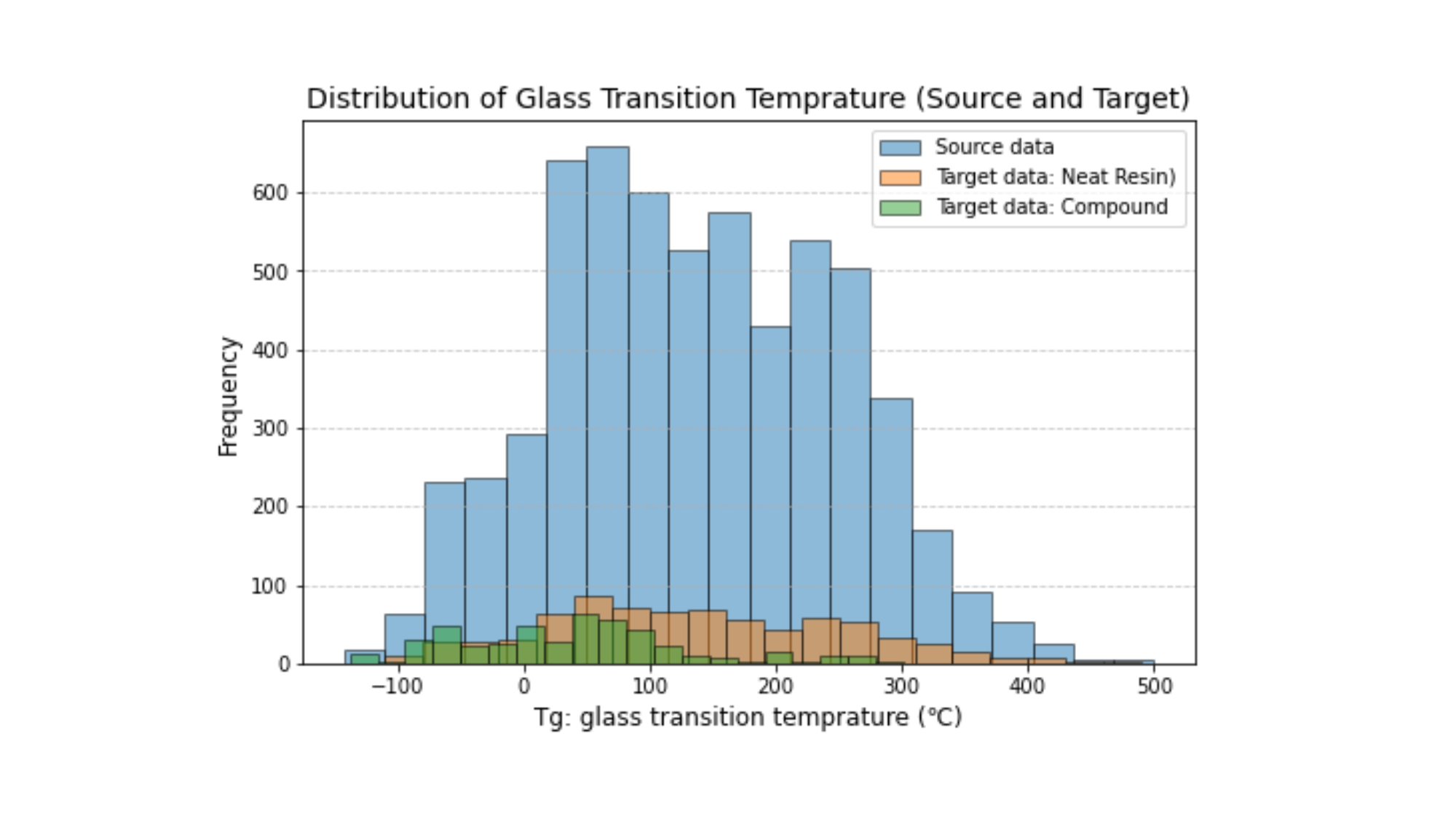}
 \end{tabular}
 \caption{The distribution of glass transition temperature. The area of two target data: Neat Resin and Compound are overlapped with the are of Source data. 
 \label{fig:tg_distribution}
 }
 \end{center}
\end{figure}

NR datasets were split into train of source data (80\%: 6003 datasets), val of source data (10\%: 750 datasets) and target data (10\%: 750 datasets). These train of source data and val of source data of NR datasets were used as building pretrained model, and target data of NR datasets and Comp datasets were split into \red{train/val data (60\%) and test data (40\%).\red{The val data was used for hyperparameter tuning by Bayesian optimization.} The Tg values (${}^\circ$C) of these datasets is distributed as Figure~\ref{fig:tg_distribution} }. 

\subsection{feature selection}
Our proposal model has the advantage of flexibility and is highly extended for various data formats. Meanwhile, feature selection is optionally necessary because features from the pretrained model play a role in feature augmentation and bring enormous features. In this study, we selected RDKit features when target datasets are Comp datasets, which particularly has huge features. The feature selection was operated by Sequential Features Selection(SFS); forward direction, $R^2$ scoring and five cross validation. The feature importance of NR test datasets was used for this feature selection.

\subsection{Evaluation}
We evaluated our model by using two target datasets: NR datasets of target and Comp datasets. As a benchmark, the model using RDKit descriptors were built up. For prediction model, Gradient Boosting and Random Forest were used in both Method~1 and 2. For target data division, we divided each target \red{train/val} datasets into the appropriate ratio. 
\red{The $R^2$ and $RMSE$ values of target test data, which was picked up before training at 40\% from target datasets, was used as an indicator.} 
\red{Hyperparameter tuning for the Gradient Boosting/Random Forest was performed by Bayesian optimization implemented in Optuna framework~\cite{akiba2019optuna}, optimizing the cross-validation $R^2$ score. The search space included n estimators [50,200], learning rate [0.01,0.2], max depth [2,8], min samples split [2,8], and min samples leaf [1,4]. A total of 50 trials were conducted, and the best configuration was selected based on the highest mean $R^2$ score from three-fold cross-validation.}

\section{Results and Discussion}
\label{headings}
\subsection{pretrained model}
The GCN was trained with GPU using approximately 6,000 source data points, with 750 data points utilized for validation purposes.
the validation loss reached a plateau at approximately 500 epochs, early stopping was employed to terminate the training process.

\subsection{transferred to target data of Neat Resin datasets}\label{subsec:experiments_nr}

\begin{figure}[t]
\centering
  \includegraphics[clip, width=0.96\columnwidth]
  {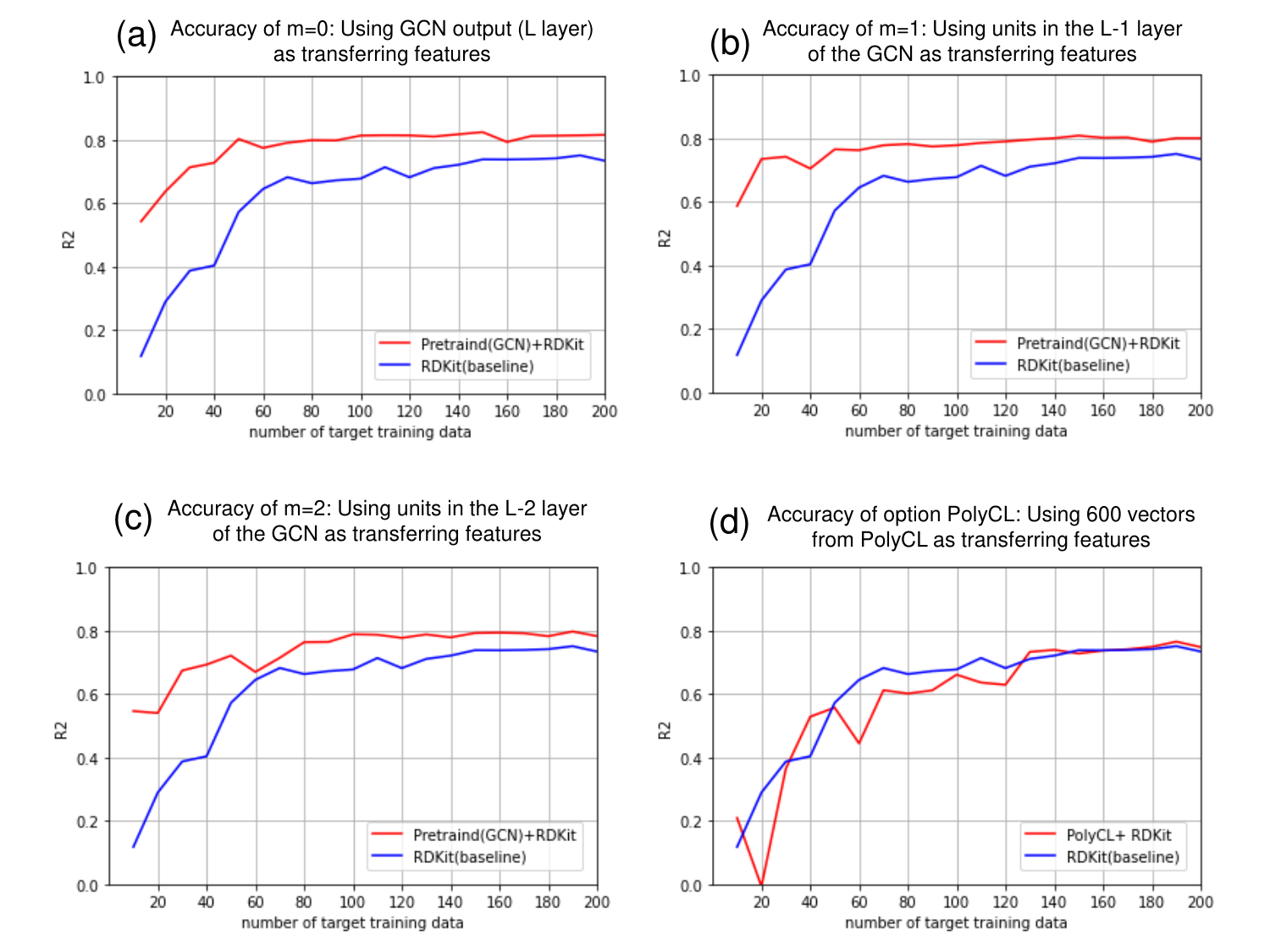}
 \caption{Results of transferred to Neat Resin
 Results of transferred to target Neat resin datasets. In each plot, the
horizontal axis represents the number of training data points in the target domain,
while the vertical axis shows the R2 values of the resulting models. It can be observed
that feature transfer from the GNN significantly improves the prediction accuracy of
polymer properties across all cases. Among the three options, feature transfer from the
L - 1-th layer of the GNN model (the case (b)) exhibited the highest transfer efficiency.}
 \label{fig:results_r2_NR}
\end{figure}

The results of the target NR data are presented in Figure ~\ref{fig:results_r2_NR}. Figure~\ref{fig:results_r2_NR} illustrates the test R² scores of the model using the test data from the NR dataset, while Appendix~\ref{sec:appendix_a} shows the RMSE scores for the test data.

As shown in Figure~\ref{fig:results_r2_NR}~(a), in the case of Option 1 (feature transfer from the final layer of the GNN model), the test R² score exceeded 0.7 with only \red{30 training data points and reached approximately 0.8 using 50 to 60 training data points. In contrast, the benchmark R² score was around 0.48 at 30 training data points. The benchmark scores surpassed 0.6 at 60 training data points, but at least 140 training data points were required to exceed 0.7.} Even with 200 training data points, the benchmark R² score remained below 0.8. These results indicate that our model required only \red{21\%} of the training data needed by the benchmark to achieve an accuracy of 0.7.

As depicted in Figure~\ref{fig:results_r2_NR}~(b), in the case of Option 2 (feature transfer from the (L - 1)-th layer of the GNN model), the test R² score reached \red{0.59} with just 10 training data points and exceeded 0.7 with 20 training data points. Similar to Option 1, if an R² score of 0.7 is considered the benchmark, Option 2 achieved the same accuracy obtained by the benchmark with \red{140} training data points using only 20 data points, which corresponds to just \red{14\%} of the required training data. These findings suggest that Option 2 is a more effective approach than Option 1. The difference between Option 1 and 2 can be attributed to the nature of the transferred features. Option 1 transfers features that have been transformed by the GNN model into scalar values for each vertex, whereas Option 2 transfers features extracted from an intermediate layer of the GNN. These intermediate features are likely to offer greater generality as feature representations compared to the final-layer outputs.

As illustrated in Figure~\ref{fig:results_r2_NR}~(c), in the case of Option 3 (feature transfer from the (L - 2)-th layer of the GNN model), the test R² score, despite some fluctuations, reached approximately \red{0.7 using 30 to 60} training data points. While Option 3 performed slightly worse than Option 2 when fewer than 80 training data points were used, it consistently outperformed the benchmark, \red{both for cases with fewer than 100 training data points and at around 200 training data points. As shown in Figure~\ref{fig:results_r2_NR}~(d), in the case of Option PolyCL (feature transfer of 600-dimensional features obtained through PolyCL), the results fluctuated, but no significant improvement over the benchmark was observed.}

The RMSE scores followed a similar trend to the R² scores, and the same pattern was observed with Random Forest. The RMSE results are presented in Appendix A5 and A6, while the results for Random Forest are provided in Appendix~\ref{sec:appendix_a}. Detailed results and hyperparameter settings are given in Appendix~\ref{sec:appendix_d}.

These findings demonstrate that Option 1, 2, and 3 consistently outperform the benchmark across all cases. As discussed in Figure~\ref{fig:results_r2_NR}~(b), these results suggest that Option 1, 2, and 3 adopt different feature transfer approaches, utilizing features with distinct properties. Furthermore, comparing Option 2 and 3 reveals that Option 2 provides greater performance improvements relative to the number of training data points in the target domain. This observation suggests that the optimal choice of which layer’s features to transfer should be determined based on the specific requirements of each task.

\red{Additionally, our results confirm that Option PolyCL (feature transfer of 600-dimensional features obtained through PolyCL) can be easily incorporated into our model. While the current results do not yet achieve high accuracy, we believe that with further refinements in the integration process and advancements in pre-trained models, this approach has the potential to become an effective method.As an example of interpretability analysis, the top 20 important features identified through Sequential Feature Selection (SFS) are presented in Appendix~\ref{sec:appendix_b}. In addition, a table of the metric values is provided in Appendix~\ref{sec:appendix_c}.}

\subsection{transferred to target data of Compounding datasets}\label{subsec:experiments_compounding}

\begin{figure}[t]
 \begin{center}
 \begin{tabular}{c}
  \includegraphics[clip, width=0.96\columnwidth]
  {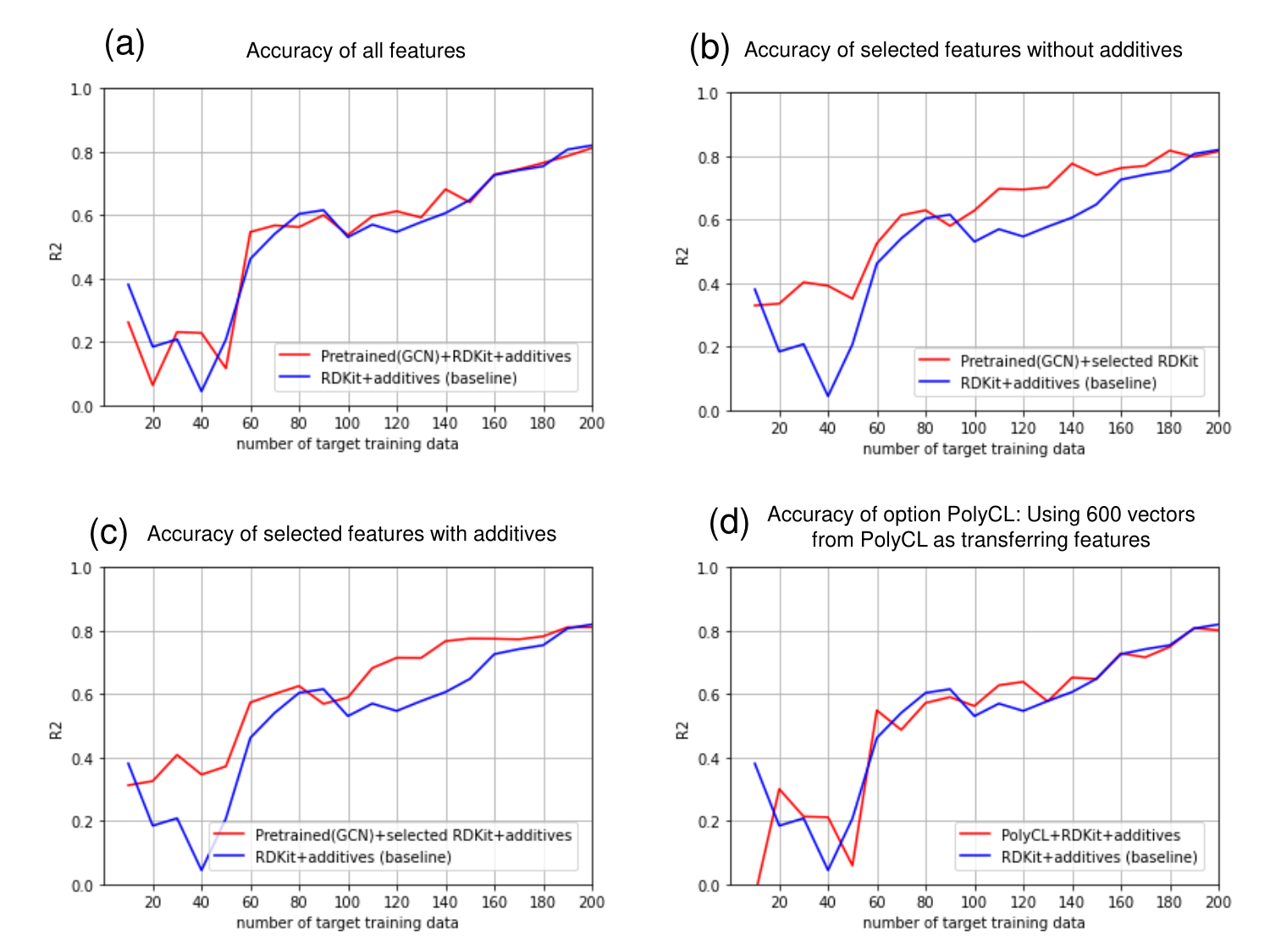}
 \end{tabular}
 \caption{Results of transferred to target compound datasets
 left top: (a) all features, right top: (b) selected RDKit + GCN features, left down: (c) selected RDKit + GCN features + additives, right down: (d) RDKit + features from PolyCL. 
 In each plot, the horizontal axis represents the number of training data points in the target domain, while the vertical axis shows the $R^2$ values of the resulting models. 
 \label{fig:results_r2_Comp}
 }
 \end{center}
\end{figure}

The results for the target Comp dataset are presented in Figure~\ref{fig:results_r2_Comp}, illustrating the outcomes under different feature selection conditions. All results in Figure~\ref{fig:results_r2_Comp}~(a)–(c) correspond to feature transfer from the L-1 layer. Figure~\ref{fig:results_r2_Comp} depicts the R² scores obtained using Gradient Boosting. \red{For RMSE, the results are shown in Appendix~\ref{sec:appendix_a}, while the R² and RMSE scores for Random Forest are provided in Appendix~\ref{sec:appendix_a}.
}
First, Figure~\ref{fig:results_r2_Comp}~(a) presents the results when all features are utilized. \red{In cases where the training dataset comprises fewer than 100 samples, negative transfer frequently occurs. Conversely, when the training dataset exceeds 100 samples, the transfer effect becomes either neutral or positive.}

Figure~\ref{fig:results_r2_Comp}~(b) displays the results when the RDKit descriptors are restricted to the 10 most important features and combined with features from the GCN model. While the model slightly underperforms the benchmark when the number of training samples is \red{10 or 90, positive transfer is generally observed. Furthermore, when the number of training samples reaches 110, the test R² score exceeds 0.7. Given that the benchmark requires 160 samples to achieve a similar level of accuracy, this result implies that our model reduces the required training data by 69\% to reach the benchmark performance.}
Figure~\ref{fig:results_r2_Comp}~(c) shows the results when additional encoded features related to additives are incorporated. The observed trends are consistent with those in Figure~\ref{fig:results_r2_Comp}~(b).The table of the metric values is provided in Appendix~\ref{sec:appendix_c}.As supporting experiments, the impacts of hyperparameter tuning illustrate in Appendix~\ref{sec:appendix_e}.

These empirical evidences support the notion that appropriate feature selection facilitates positive transfer.
In this study, feature selection was applied to the RDKit descriptors, and additives information was encoded using one-hot vectors. \red{However, further improvements in accuracy are anticipated through the development of methods that more densely encode additive information, such as leveraging encoding techniques from large language models (LLMs).}

Furthermore, Figure~\ref{fig:results_r2_Comp}~(d) presents the results for the optional PolyCL method, which employs 600-dimensional feature transfer obtained via PolyCL. No clear superiority over the benchmark was observed in this case. However, as discussed in Section 4.2, leveraging self-supervised learning approaches like PolyCL as pre-trained models is considered reasonable and holds potential. Thus, further study need be to be done as an anticipated future work.

\subsection{Discussions for model interpretability}\label{subsec:importance_interpretability}
Understanding the features that contribute to the predictions not only serves to validate the model’s reliability but also provides more intuitive guidance for polymer design. 
In the ensemble machine learning model employed in this study, feature importance can be calculated, allowing for the estimation of influential features based on their ranking with respect to the prediction outcomes.

However, it is important to note that in this study, a complex nonlinear feature extractor (GCN) is used to derive features from SMILES, making it impossible to directly associate the obtained feature importance with individual elements of the raw data (this point is also discussed in Appendix~\ref{sec:appendix_b}). 
The use of such an uninterpretable model imposes a limitation in that, when viewing the raw data as input and the prediction as output, one cannot necessarily assume a causal relationship between the two. 
From a broader perspective, it also means that the resulting predictive model cannot be readily interpreted as a good surrogate for the actual physical phenomena.

One possible approach to addressing this limitation is to incorporate interpretability into the feature extractor. 
For example, by using a transformer-based encoder with a graph attention mechanism instead of a GCN, it becomes possible to partially visualize the relationships between the learned representations and the raw input data. 
By combining this information with the previously discussed feature importance, it may be possible to construct a model that allows for a certain degree of interpretability from the raw data to the prediction outcomes.

\section{Conclusion}
\label{others}
This paper presents a multi-modal cascade model with feature transfer as a methodology for adaptive experimental design, with the objective of modifying the characteristics of polymer property prediction.  In several trials using experimental polymer datasets as a proof of concept, it was demonstrated that the proposed methodology exhibited superior accuracy compared to the benchmark. 

\red{In the optimal configuration, satisfactory performance was achieved using only 20 training data points, accounting for merely 14\% of the total training dataset. This contrasts markedly with the benchmark model, which requires a substantially larger number of training samples, specifically 140, to attain comparable results.}

It was also proved that positive transfer can be obtained with appropriate feature selection. 
Subsequent research will see the model expanded to encompass the intricacies of polymer composition, process condition, and the nuances of their physical properties. 
In particular, efforts will be directed toward developing a refined pre-trained model capable of predicting multiple properties and designing a method to encode additive information into dense vectors. 
Regarding pre-trained models, this study primarily employed GCN. However, future research will explore the application of other pre-trained models, such as Transformer-based models, and further develop them both internally and externally. 
In this study, the predicted property was the glass transition temperature (Tg), however, future work is expected to predict the properties that are more meaningful from an engineering perspective, such as electrical and mechanical properties.
Additionally, for encoding additive information, the utilization of LLMs is expected to enable not only the encoding of additive names but also their meanings and functionalities.

\section*{Acknowledgments and Disclosure of Funding}
PoLyInfo datasets was provided from NIMS (National Institute of Material Science) by following contract with them.

This work was supported by Grants-in-Aid from the Japan Society for the Promotion of Science(JSPS) for Scientific Research(KAKENHI grant nos. JP20K19871, JP23K28146, JP25K03086 and JP24K20836 to K.M.).

\medskip

\bibliographystyle{unsrt}
\bibliography{ref}

\clearpage

\appendix

\section{results of transferred to target data}\label{sec:appendix_a}

Following figures illustrate the behavior of each metric ($R^2$ or $RMSE$) . These relate to Section~\ref{subsec:experiments_nr} and Section~\ref{subsec:experiments_compounding}. 
\begin{figure}[h]
 \begin{center}
 \begin{tabular}{c}
  \includegraphics[clip, width=1.0\columnwidth]
  {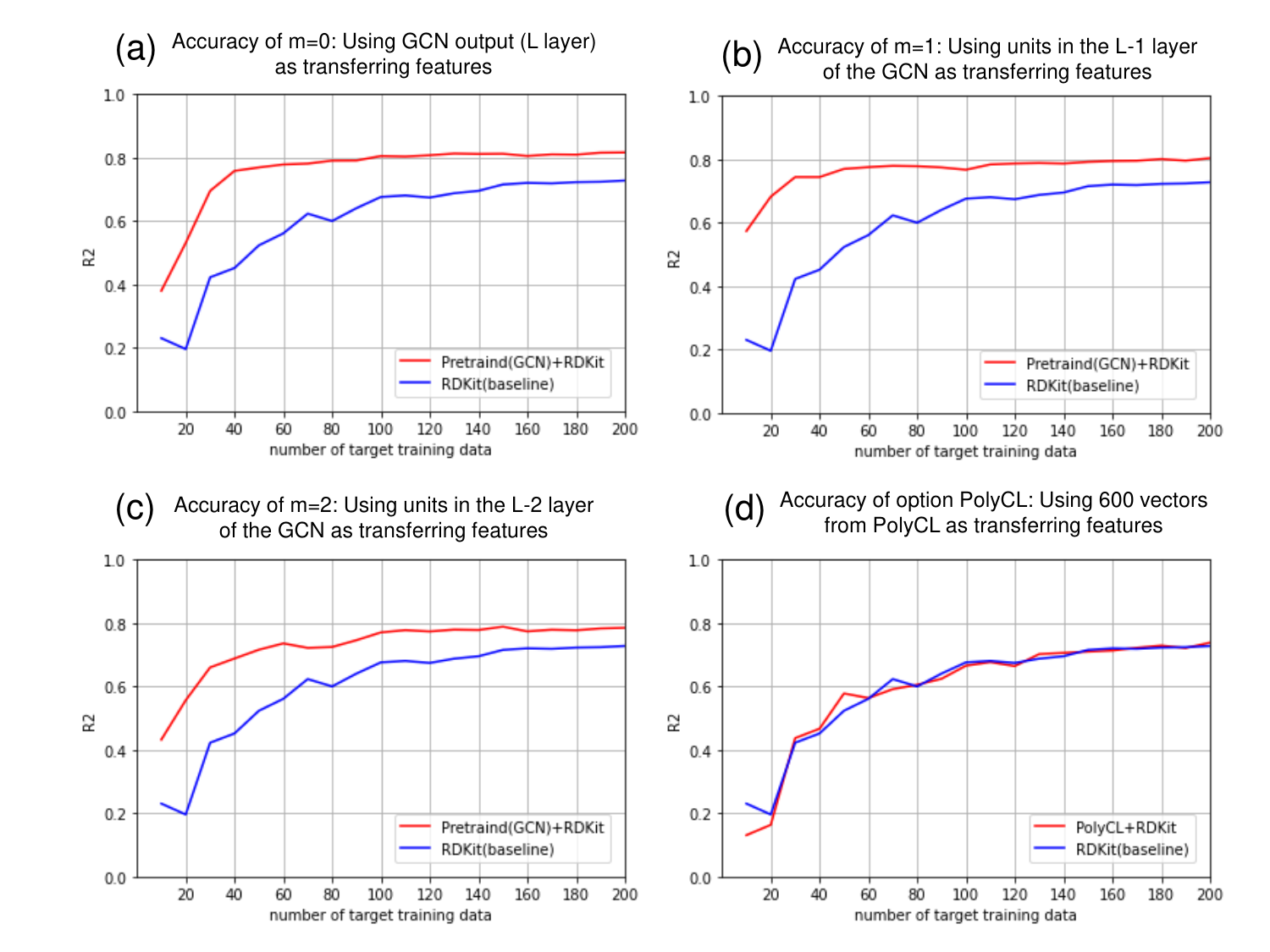}
 \end{tabular}
 \caption{Random Forest $R^2$ score of transferred to target neat resin datasets.
 left top: (a) Option1(m=0), right top: (b) Option2(m=1), left down: (c) Option3(m=2), right down: (d) Option PolyCL 
 \label{fig/mlst_A1}
 }
 \end{center}
\end{figure}
%
\clearpage
\begin{figure}[h]
 \begin{center}
 \begin{tabular}{c}
  \includegraphics[clip, width=1.0\columnwidth]
  {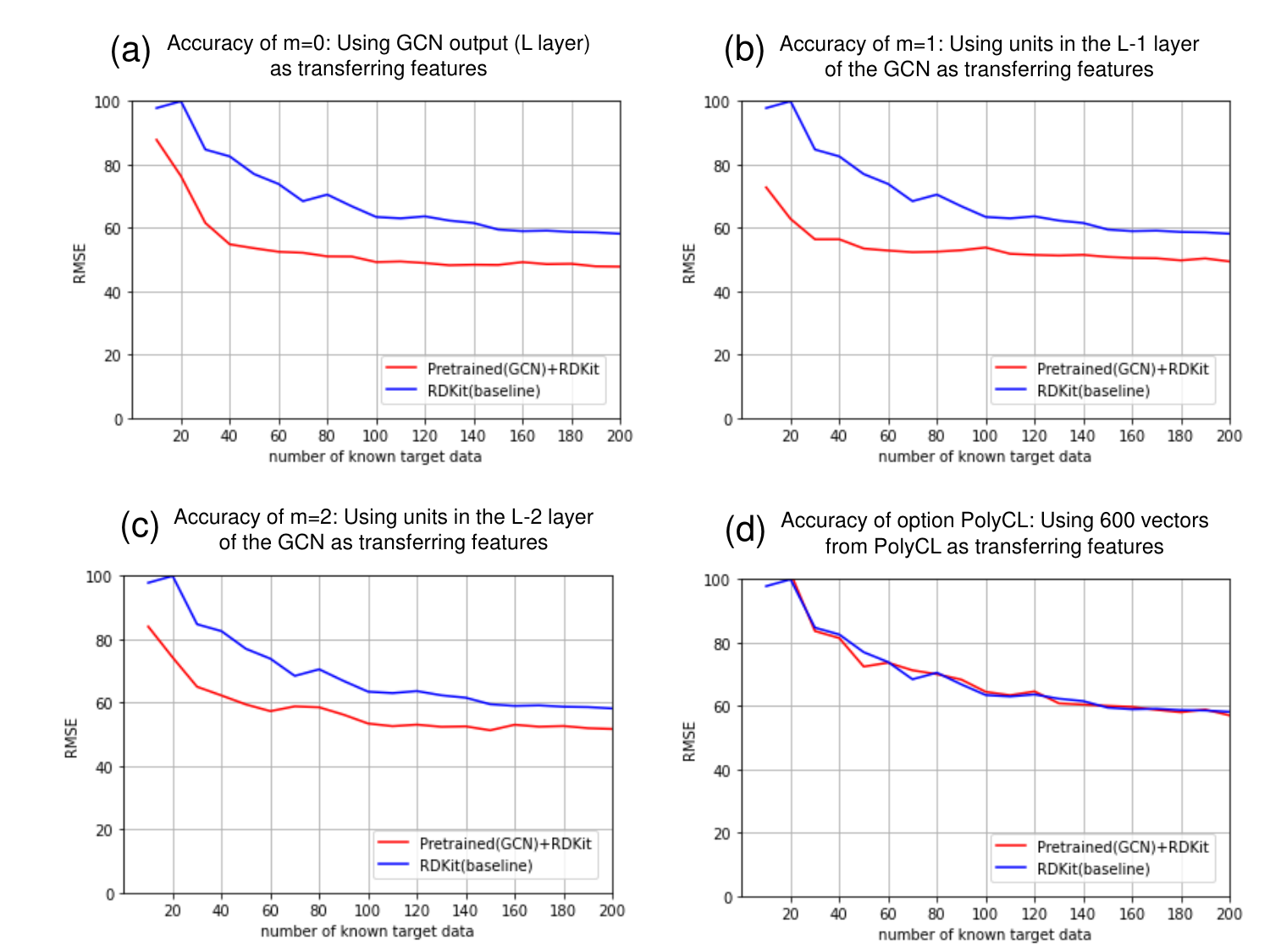}
 \end{tabular}
 \caption{Random Forest $RMSE$ score of transferred to target neat resin datasets.
 left top: (a) Option1(m=0), right top: (b) Option2(m=1), left down: (c) Option3(m=2), right down: (d) Option PolyCL 
 \label{fig/mlst_A2}
 }
 \end{center}
\end{figure}
%
\clearpage
\begin{figure}[h]
 \begin{center}
 \begin{tabular}{c}
  \includegraphics[clip, width=1.0\columnwidth]
  {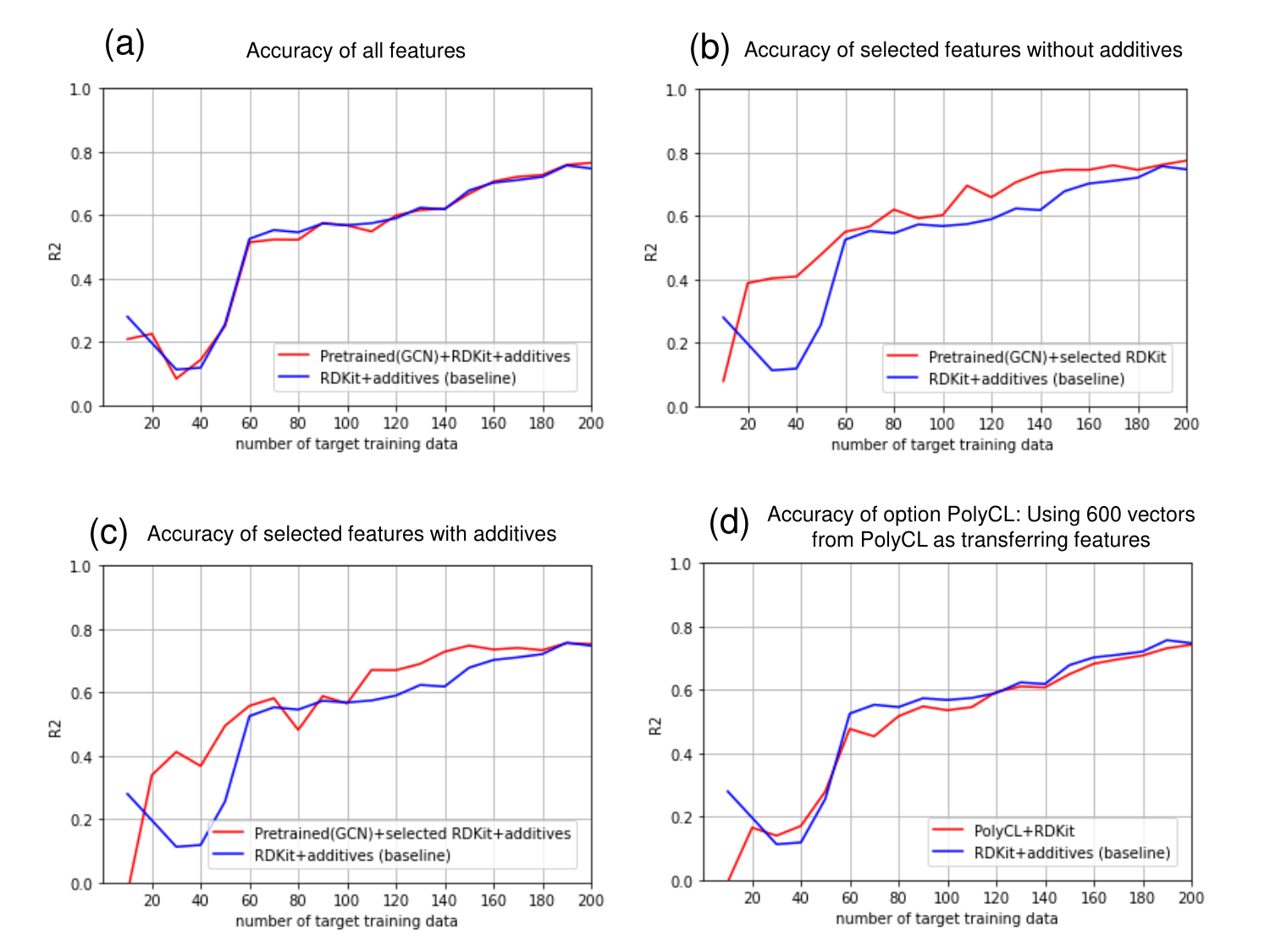}
 \end{tabular}
 \caption{Random Forest $R^2$ score of transferred to target compound datasets.
 left top: (a) all features, right top: (b) selected RDKit + GCN features, left down: (c) selected RDKit + GCN features + additives, right down: (d) RDKit + features from PolyCL
 \label{fig/mlst_A3}
 }
 \end{center}
\end{figure}
%
\clearpage
\begin{figure}[h]
 \begin{center}
 \begin{tabular}{c}
  \includegraphics[clip, width=1.0\columnwidth]
  {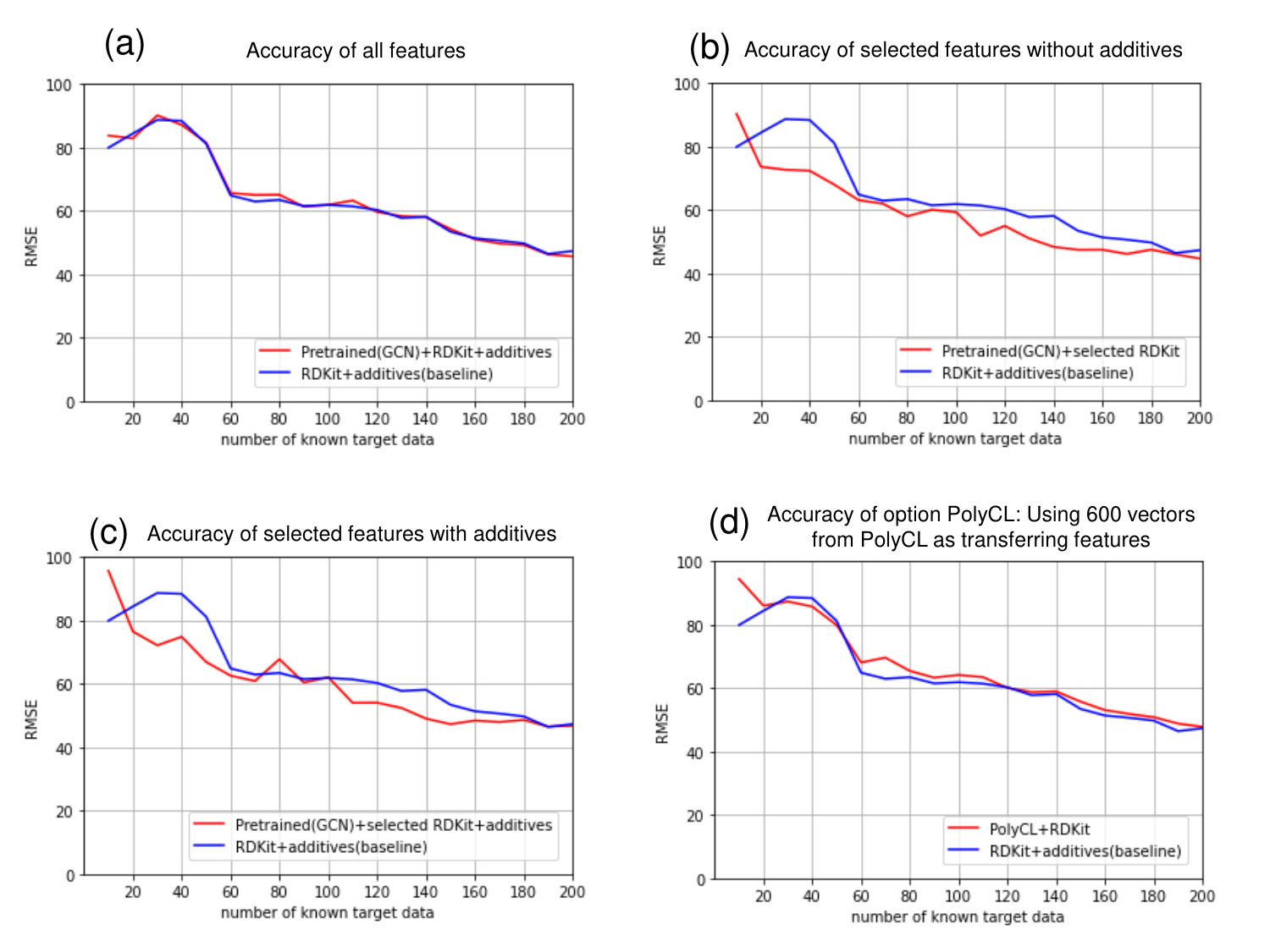}
 \end{tabular}
 \caption{Random Forest $R^2$ score of transferred to target compound datasets.
left top: (a) all features, right top: (b) selected RDKit + GCN features, left down: (c) selected RDKit + GCN features + additives, right down: (d) RDKit + features from PolyCL 
 \label{fig/mlst_A4}
 }
 \end{center}
\end{figure}
%
\clearpage
\begin{figure}[h]
 \begin{center}
 \begin{tabular}{c}
  \includegraphics[clip, width=1.0\columnwidth]
  {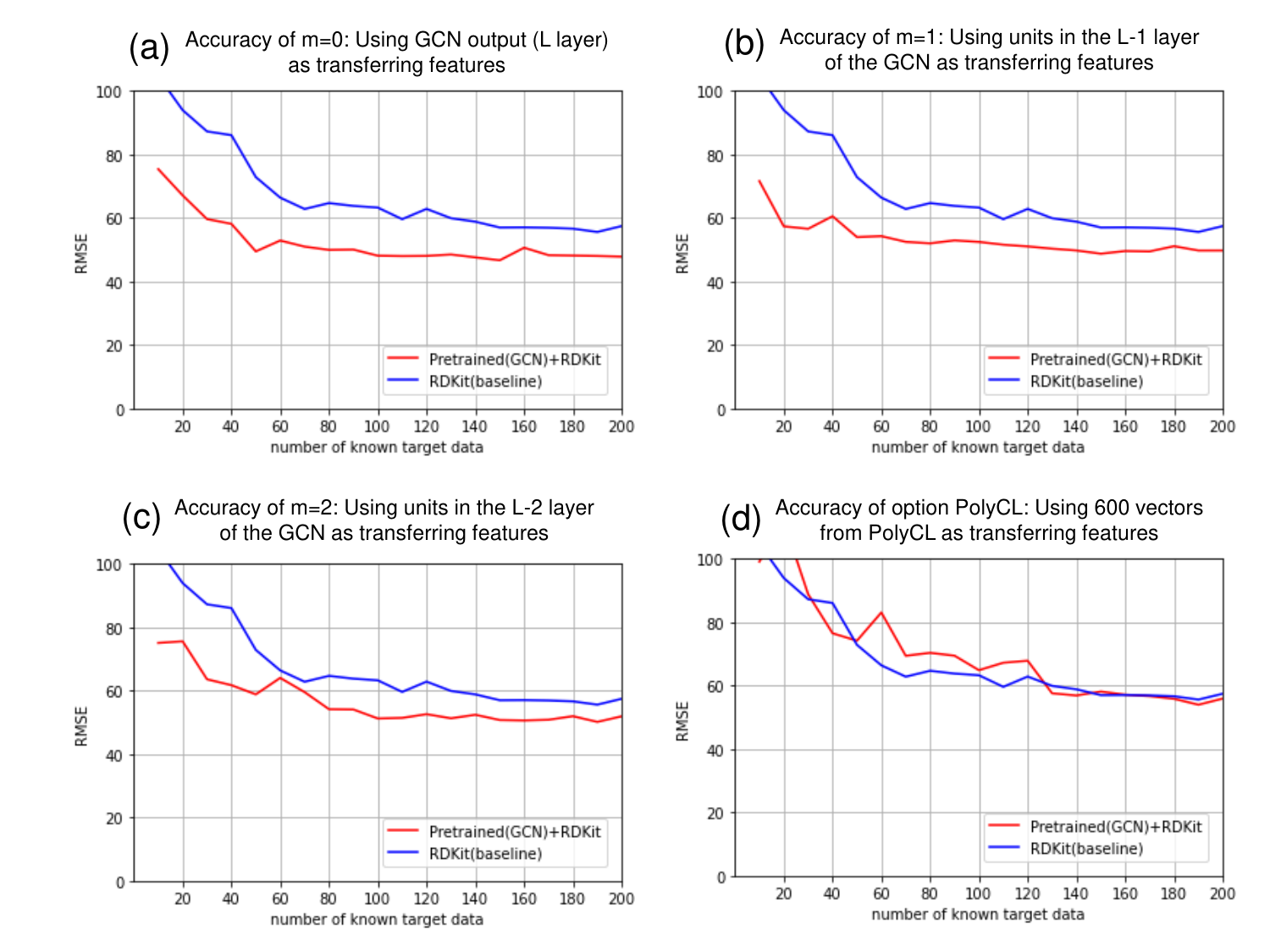}
 \end{tabular}
 \caption{Gradient Boosting $RMSE$ score of transferred to target Neat Resin datasets.
left top: (a) Option1(m=0), right top: (b) Option2(m=1), left down: (c) Option3(m=2), right down: (d) Option PolyCL
 \label{fig/mlst_A5}
 }
 \end{center}
\end{figure}
%
\clearpage
\begin{figure}[h]
 \begin{center}
 \begin{tabular}{c}
  \includegraphics[clip, width=1.0\columnwidth]
  {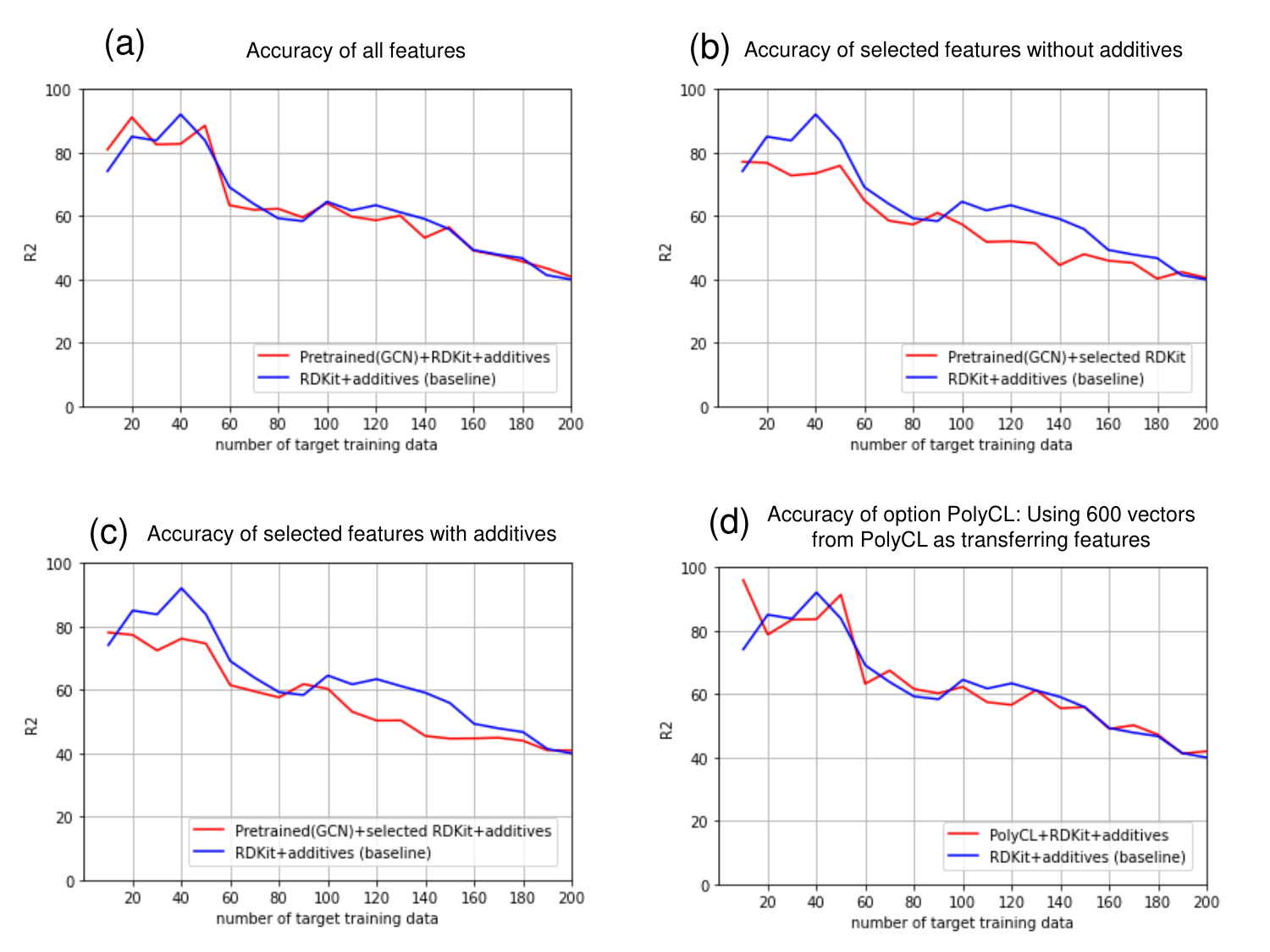}
 \end{tabular}
 \caption{Gradient Boosting $RMSE$ score of transferred to target compound datasets.
left top: (a) all features, right top: (b) selected RDKit + GCN features, left down: (c) selected RDKit + GCN features + additives, right down: (d) RDKit + features from PolyCL 
 \label{fig/mlst_A6}
 }
 \end{center}
\end{figure}
%

\clearpage
\section{Feature Importance}\label{sec:appendix_b}
Following figure illustrates feature importance of neat resin option1. This related Section~\ref{subsec:experiments_nr}. 
\begin{figure}[h]
 \begin{center}
 \begin{tabular}{c}
  \includegraphics[clip, width=1.0\columnwidth]
  {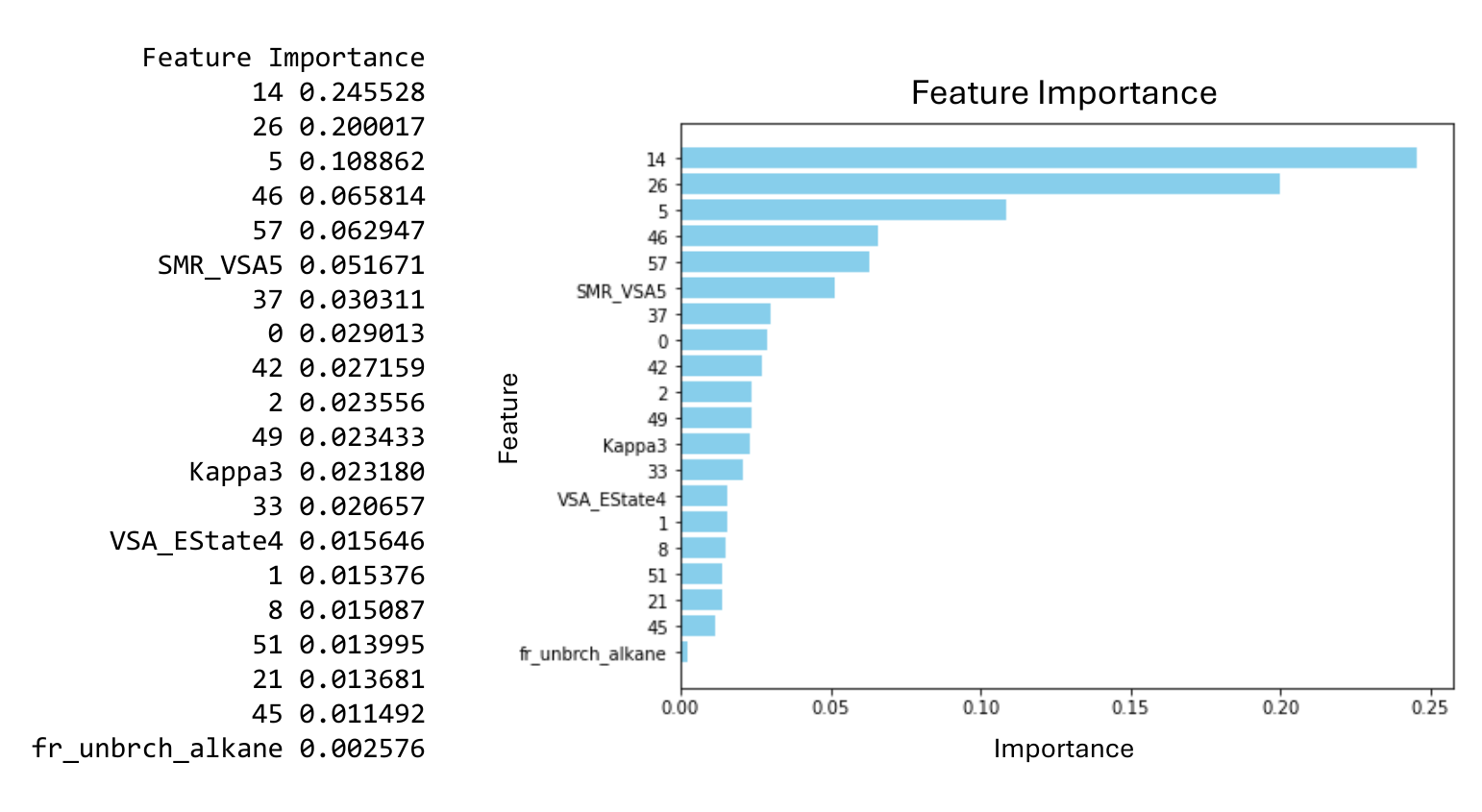}
 \end{tabular}
 \caption{Feature Importance of neat resin option1. We can observed that features from GCN such as feature 14,26,5,46 etc... are mixed.In this study, the predictive model is constructed by transforming the original features into feature vectors using a deep learning model. Therefore, it is important to note that when examining feature contributions, the interpretability of the original feature importance cannot be directly inferred.
 \label{fig/mlst_B1_revised.pdf}
 }
 \end{center}
\end{figure}

\clearpage 
\section{results of $R^2$ scores}\label{sec:appendix_c}
Following Tables illustrate $R^2$ score of each model in 10,40,70,100,130,160,190 training data. The names of the models abbreviated in the table are as follows. Option1:Using GCN output(L layer) as transferring features, Option2:Using units in the L-1 layer of the GCN as transferring features, Option3:Using units in the L-2 layer of the GCN as transferring features, PolyCL:Using 600 vectors from PolyCL as transferring features, FS1(feature selection1):selected RDKit + GCN features, FS2(feature selection2):selected RDKit + GCN features + additives.
\begin{itemize}
    \item Table~\ref{tab:label_of_C1} shows the result of $R^2$ values obtained when gradient boosting was applied to the Neat Resin datasets.
    \item Table~\ref{tab:label_of_C2} shows the result of $RMSE$ values obtained when gradient boosting was applied to the Neat Resin datasets.
    \item Table~\ref{tab:label_of_C3} shows the result of $R^2$ values obtained when gradient boosting was applied to the Compound datasets.
    \item Table~\ref{tab:label_of_C4} shows the result of $RMSE$ values obtained when gradient boosting was applied to the Compound datasets.
    \item Table~\ref{tab:label_of_C5} shows the result of $R^2$ values obtained when random forest was applied to the Neat Resin datasets. 
    \item Table~\ref{tab:label_of_C6} shows the result of $RMSE$ values obtained when random forest was applied to the Neat Resin datasets. 
    \item Table~\ref{tab:label_of_C7} shows the result of $R^2$ values obtained when random forest was applied to the compound datasets. 
    \item Table~\ref{tab:label_of_C8} shows the result of $RMSE$ values obtained when random forest was applied to the compound datasets. 
\end{itemize}
\newpage
\begin{table}[t]
    \centering
    \caption{$R^2$ score of neat resin by gradient boosting}
    \label{tab:label_of_C1}
    \begin{tabular}{lrrrrrrr}
\toprule
$N_{\mathrm{train}}$ & 10 & 40 & 70 & 100 & 130 & 160 & 190 \\
\midrule
baseline & 0.118 & 0.404 & 0.682 & 0.678 & 0.711 & 0.738 & 0.751 \\
Option1 & 0.543 & 0.728 & 0.791 & 0.814 & 0.811 & 0.794 & 0.814 \\
Option2 & 0.587 & 0.705 & 0.778 & 0.779 & 0.796 & 0.802 & 0.801 \\
Option3 & 0.546 & 0.693 & 0.715 & 0.789 & 0.788 & 0.794 & 0.798 \\
PolyCL & 0.209 & 0.529 & 0.612 & 0.661 & 0.733 & 0.737 & 0.766 \\
\bottomrule
\end{tabular}
\end{table}
\begin{table}[t]
    \centering
    \caption{$RMSE$ score of neat resin by gradient boosting}
    \label{tab:label_of_C2}
    \begin{tabular}{lrrrrrrr}
\toprule
$N_{\mathrm{train}}$ & 10 & 40 & 70 & 100 & 130 & 160 & 190 \\
\midrule
baseline & 104.713 & 86.089 & 62.826 & 63.278 & 59.927 & 57.047 & 55.602 \\
Option1 & 75.394 & 58.151 & 50.978 & 48.125 & 48.503 & 50.633 & 48.035 \\
Option2 & 71.625 & 60.551 & 52.485 & 52.465 & 50.337 & 49.592 & 49.752 \\
Option3 & 75.085 & 61.747 & 59.557 & 51.231 & 51.287 & 50.605 & 50.154 \\
PolyCL & 99.137 & 76.529 & 69.422 & 64.874 & 57.577 & 57.153 & 53.971 \\
\bottomrule
\end{tabular}
\end{table}
\begin{table}[t]
    \centering
    \caption{$R^2$ score of compound by gradient boosting}
    \label{tab:label_of_C3}
    \begin{tabular}{lrrrrrrr}
\toprule
$N_{\mathrm{train}}$ & 10 & 40 & 70 & 100 & 130 & 160 & 190 \\
\midrule
baseline & 0.381 & 0.044 & 0.541 & 0.531 & 0.578 & 0.726 & 0.807 \\
all\_features & 0.262 & 0.228 & 0.568 & 0.537 & 0.593 & 0.729 & 0.787 \\
FS1 & 0.330 & 0.392 & 0.614 & 0.629 & 0.702 & 0.763 & 0.798 \\
FS2 & 0.312 & 0.346 & 0.601 & 0.590 & 0.714 & 0.775 & 0.811 \\
PolyCL & -0.039 & 0.211 & 0.487 & 0.562 & 0.577 & 0.729 & 0.808 \\
\bottomrule
\end{tabular}
\end{table}
\begin{table}[t]
    \centering
    \caption{$RMSE$ score of compound by gradient boosting}
    \label{tab:label_of_C4}
    \begin{tabular}{lrrrrrrr}
\toprule
$N_{\mathrm{train}}$ & 10 & 40 & 70 & 100 & 130 & 160 & 190 \\
\midrule
baseline & 74.114 & 92.093 & 63.822 & 64.528 & 61.184 & 49.288 & 41.357 \\
all\_features & 80.932 & 82.750 & 61.939 & 64.051 & 60.119 & 49.071 & 43.483 \\
FS1 & 77.097 & 73.451 & 58.519 & 57.377 & 51.379 & 45.893 & 42.327 \\
FS2 & 78.099 & 76.170 & 59.516 & 60.336 & 50.355 & 44.687 & 40.937 \\
PolyCL & 95.986 & 83.636 & 67.458 & 62.299 & 61.249 & 49.052 & 41.232 \\
\bottomrule
\end{tabular}
\end{table}
\begin{table}[t]
    \centering
    \caption{$R^2$ score of neat resin by random forest}
    \label{tab:label_of_C5}
    \begin{tabular}{lrrrrrrr}
\toprule
$N_{\mathrm{train}}$ & 10 & 40 & 70 & 100 & 130 & 160 & 190 \\
\midrule
baseline & 0.230 & 0.452 & 0.623 & 0.676 & 0.688 & 0.720 & 0.724 \\
Option1 & 0.380 & 0.758 & 0.781 & 0.805 & 0.813 & 0.805 & 0.816 \\
Option2 & 0.574 & 0.744 & 0.780 & 0.767 & 0.789 & 0.795 & 0.796 \\
Option3 & 0.432 & 0.688 & 0.722 & 0.771 & 0.780 & 0.774 & 0.783 \\
PolyCL & 0.131 & 0.467 & 0.591 & 0.666 & 0.702 & 0.713 & 0.721 \\
\bottomrule
\end{tabular}
\end{table}
\begin{table}[t]
    \centering
    \caption{$RMSE$ score of neat resin by random forest}
    \label{tab:label_of_C6}
    \begin{tabular}{lrrrrrrr}
\toprule
$N_{\mathrm{train}}$ & 10 & 40 & 70 & 100 & 130 & 160 & 190 \\
\midrule
baseline & 97.802 & 82.541 & 68.414 & 63.439 & 62.290 & 58.948 & 58.563 \\
Option1 & 87.786 & 54.796 & 52.126 & 49.199 & 48.196 & 49.174 & 47.854 \\
Option2 & 72.784 & 56.386 & 52.320 & 53.776 & 51.260 & 50.489 & 50.347 \\
Option3 & 84.001 & 62.236 & 58.795 & 53.383 & 52.347 & 53.005 & 51.913 \\
PolyCL & 103.939 & 81.413 & 71.253 & 64.469 & 60.826 & 59.686 & 58.861 \\
\bottomrule
\end{tabular}
\end{table}
\begin{table}[t]
    \centering
    \caption{$R^2$ score of compound by random forest}
    \label{tab:label_of_C7}
    \begin{tabular}{lrrrrrrr}
\toprule
$N_{\mathrm{train}}$ & 10 & 40 & 70 & 100 & 130 & 160 & 190 \\
\midrule
baseline & 0.280 & 0.118 & 0.553 & 0.568 & 0.624 & 0.702 & 0.757 \\
all\_features & 0.209 & 0.144 & 0.523 & 0.568 & 0.616 & 0.706 & 0.759 \\
FS1 & 0.079 & 0.409 & 0.566 & 0.603 & 0.706 & 0.746 & 0.761 \\
FS2 & -0.033 & 0.368 & 0.582 & 0.565 & 0.690 & 0.735 & 0.756 \\
PolyCL & -0.006 & 0.171 & 0.453 & 0.536 & 0.611 & 0.682 & 0.731 \\
\bottomrule
\end{tabular}
\end{table}
\begin{table}[t]
    \centering
    \caption{$RMSE$ score of compound by random forest}
    \label{tab:label_of_C8}
    \begin{tabular}{lrrrrrrr}
\toprule
$N_{\mathrm{train}}$ & 10 & 40 & 70 & 100 & 130 & 160 & 190 \\
\midrule
baseline & 79.940 & 88.430 & 62.966 & 61.888 & 57.782 & 51.380 & 46.440 \\
all\_features & 83.789 & 87.163 & 65.055 & 61.937 & 58.359 & 51.044 & 46.244 \\
FS1 & 90.368 & 72.417 & 62.018 & 59.369 & 51.051 & 47.491 & 46.007 \\
FS2 & 95.713 & 74.898 & 60.914 & 62.120 & 52.428 & 48.443 & 46.552 \\
PolyCL & 94.485 & 85.769 & 69.638 & 64.176 & 58.762 & 53.077 & 48.817 \\
\bottomrule
\end{tabular}
\end{table}

\clearpage
\section{hyperparameter information}\label{sec:appendix_d}
Following Tables illustrate hyper-parameters in 10,40,70,100,130,160,190 training data. These values are calculated by Bayesian optimization.
\begin{itemize}
    \item Table~\ref{tab:label_of_D1-1} to Table~\ref{tab:label_of_D1-5} shows the hyperparameter that optimized when gradient boosting was applied to the Neat Resin datasets.
    \item Table~\ref{tab:label_of_D2-1} to Table~\ref{tab:label_of_D2-5} shows the hyperparameter that optimized when gradient boosting was applied to the compound datasets.
    \item Table~\ref{tab:label_of_D3-1} to Table~\ref{tab:label_of_D3-5} shows hyperparameter that optimized when random forest was applied to the Neat Resin datasets.
    \item Table~\ref{tab:label_of_D4-1} to Table~\ref{tab:label_of_D4-5} shows hyperparameter that optimized when random forest was applied to the compound datasets.
\end{itemize}

\begin{table}[ht]
    \centering
    \caption{hyperparameter of Neat Resin, Gradient Boosting, baseline}
    \label{tab:label_of_D1-1}
    \begin{tabular}{llllllll}
\toprule
$N_{\mathrm{train}}$ & 10 & 40 & 70 & 100 & 130 & 160 & 190 \\
\midrule
n\_estimators & 151 & 70 & 94 & 199 & 102 & 179 & 187 \\
learning\_rate & 0.163 & 0.030 & 0.116 & 0.177 & 0.176 & 0.199 & 0.087 \\
max\_depth & 7 & 8 & 2 & 4 & 2 & 3 & 3 \\
min\_samples\_split & 5 & 8 & 5 & 3 & 3 & 3 & 5 \\
min\_samples\_leaf & 3 & 3 & 4 & 2 & 1 & 1 & 3 \\
\bottomrule
\end{tabular}
\end{table}
\begin{table}[ht]
    \centering
    \caption{hyperparameter of Neat Resin, Gradient Boosting, Option1}
    \label{tab:label_of_D1-2}
    \begin{tabular}{llllllll}
\toprule
$N_{\mathrm{train}}$ & 10 & 40 & 70 & 100 & 130 & 160 & 190 \\
\midrule
n\_estimators & 73 & 120 & 67 & 76 & 92 & 93 & 115 \\
learning\_rate & 0.160 & 0.149 & 0.143 & 0.072 & 0.063 & 0.046 & 0.061 \\
max\_depth & 3 & 5 & 3 & 2 & 2 & 7 & 2 \\
min\_samples\_split & 5 & 5 & 5 & 6 & 2 & 6 & 4 \\
min\_samples\_leaf & 1 & 3 & 2 & 4 & 4 & 4 & 2 \\
\bottomrule
\end{tabular}
\end{table}
\begin{table}[ht]
    \centering
    \caption{hyperparameter of Neat Resin, Gradient Boosting, Option2}
    \label{tab:label_of_D1-3}
    \begin{tabular}{llllllll}
\toprule
$N_{\mathrm{train}}$ & 10 & 40 & 70 & 100 & 130 & 160 & 190 \\
\midrule
n\_estimators & 108 & 102 & 132 & 184 & 50 & 99 & 145 \\
 learning\_rate & 0.049 & 0.029 & 0.108 & 0.193 & 0.140 & 0.053 & 0.043 \\
 max\_depth & 2 & 6 & 2 & 2 & 3 & 2 & 2 \\
 min\_samples\_split & 4 & 4 & 4 & 5 & 6 & 2 & 8 \\
 min\_samples\_leaf & 3 & 4 & 2 & 1 & 4 & 2 & 2 \\
\bottomrule
\end{tabular}
\end{table}
\begin{table}[ht]
    \centering
    \caption{hyperparameter of Neat Resin, Gradient Boosting, Option3}
    \label{tab:label_of_D1-4}
    \begin{tabular}{llllllll}
\toprule
$N_{\mathrm{train}}$ & 10 & 40 & 70 & 100 & 130 & 160 & 190 \\
\midrule
n\_estimators & 164 & 186 & 87 & 82 & 200 & 50 & 190 \\
learning\_rate & 0.087 & 0.153 & 0.072 & 0.149 & 0.136 & 0.121 & 0.095 \\
max\_depth & 6 & 5 & 3 & 2 & 2 & 2 & 3 \\
min\_samples\_split & 3 & 8 & 5 & 3 & 6 & 5 & 6 \\
min\_samples\_leaf & 2 & 4 & 3 & 3 & 2 & 1 & 3 \\
\bottomrule
\end{tabular}
\end{table}
\begin{table}[ht]
    \centering
    \caption{hyperparameter of Neat Resin, Gradient Boosting, Option PolyCL}
    \label{tab:label_of_D1-5}
    \begin{tabular}{llllllll}
\toprule
$N_{\mathrm{train}}$ & 10 & 40 & 70 & 100 & 130 & 160 & 190 \\
\midrule
n\_estimators & 59 & 130 & 192 & 150 & 50 & 93 & 185 \\
learning\_rate & 0.140 & 0.079 & 0.115 & 0.110 & 0.141 & 0.099 & 0.056 \\
max\_depth & 7 & 2 & 2 & 2 & 2 & 2 & 2 \\
min\_samples\_split & 2 & 2 & 4 & 3 & 6 & 5 & 8 \\
min\_samples\_leaf & 1 & 1 & 1 & 2 & 4 & 2 & 4 \\
\bottomrule
\end{tabular}
\end{table}
\begin{table}[ht]
    \centering
    \caption{hyperparameter of compound, Gradient Boosting, baseline}
    \label{tab:label_of_D2-1}
    \begin{tabular}{llllllll}
\toprule
$N_{\mathrm{train}}$ & 10 & 40 & 70 & 100 & 130 & 160 & 190 \\
\midrule
n\_estimators & 161 & 197 & 187 & 90 & 192 & 154 & 163 \\
learning\_rate & 0.167 & 0.010 & 0.011 & 0.038 & 0.011 & 0.109 & 0.133 \\
max\_depth & 5 & 2 & 4 & 2 & 8 & 2 & 2 \\
min\_samples\_split & 2 & 8 & 8 & 3 & 2 & 4 & 5 \\
min\_samples\_leaf & 3 & 1 & 3 & 4 & 4 & 1 & 1 \\
\bottomrule
\end{tabular}
\end{table}
\begin{table}[ht]
    \centering
    \caption{hyperparameter of compound, Gradient Boosting, all features}
    \label{tab:label_of_D2-2}
    \begin{tabular}{llllllll}
\toprule
$N_{\mathrm{train}}$ & 10 & 40 & 70 & 100 & 130 & 160 & 190 \\
\midrule
n\_estimators & 52 & 51 & 145 & 117 & 92 & 188 & 108 \\
learning\_rate & 0.076 & 0.011 & 0.129 & 0.015 & 0.063 & 0.166 & 0.097 \\
max\_depth & 7 & 8 & 2 & 8 & 5 & 4 & 5 \\
min\_samples\_split & 8 & 5 & 5 & 3 & 7 & 4 & 2 \\
min\_samples\_leaf & 4 & 2 & 2 & 1 & 3 & 1 & 3 \\
\bottomrule
\end{tabular}
\end{table}
\begin{table}[ht]
    \centering
    \caption{hyperparameter of compound, Gradient Boosting, FS1}
    \label{tab:label_of_D2-3}
    \begin{tabular}{llllllll}
\toprule
$N_{\mathrm{train}}$ & 10 & 40 & 70 & 100 & 130 & 160 & 190 \\
\midrule
n\_estimators & 147 & 179 & 193 & 129 & 50 & 95 & 79 \\
learning\_rate & 0.104 & 0.014 & 0.105 & 0.012 & 0.054 & 0.097 & 0.116 \\
max\_depth & 4 & 2 & 2 & 6 & 6 & 4 & 3 \\
min\_samples\_split & 3 & 5 & 2 & 6 & 7 & 6 & 2 \\
min\_samples\_leaf & 4 & 3 & 2 & 2 & 2 & 2 & 3 \\
\bottomrule
\end{tabular}
\end{table}
\begin{table}[ht]
    \centering
    \caption{hyperparameter of compound, Gradient Boosting, FS2}
    \label{tab:label_of_D2-4}
    \begin{tabular}{llllllll}
\toprule
$N_{\mathrm{train}}$ & 10 & 40 & 70 & 100 & 130 & 160 & 190 \\
\midrule
n\_estimators & 52 & 51 & 145 & 117 & 92 & 188 & 108 \\
learning\_rate & 0.076 & 0.011 & 0.129 & 0.015 & 0.063 & 0.166 & 0.097 \\
max\_depth & 7 & 8 & 2 & 8 & 5 & 4 & 5 \\
min\_samples\_split & 8 & 5 & 5 & 3 & 7 & 4 & 2 \\
min\_samples\_leaf & 4 & 2 & 2 & 1 & 3 & 1 & 3 \\
\bottomrule
\end{tabular}
\end{table}
\begin{table}[ht]
    \centering
    \caption{hyperparameter of compound, Gradient Boosting, Option PolyCL}
    \label{tab:label_of_D2-5}
    \begin{tabular}{llllllll}
\toprule
$N_{\mathrm{train}}$ & 10 & 40 & 70 & 100 & 130 & 160 & 190 \\
\midrule
n\_estimators & 87 & 65 & 147 & 98 & 146 & 183 & 167 \\
learning\_rate & 0.117 & 0.047 & 0.079 & 0.037 & 0.139 & 0.129 & 0.150 \\
max\_depth & 6 & 2 & 2 & 2 & 5 & 2 & 2 \\
min\_samples\_split & 4 & 7 & 7 & 7 & 4 & 3 & 5 \\
min\_samples\_leaf & 3 & 3 & 1 & 3 & 3 & 1 & 1 \\
\bottomrule
\end{tabular}
\end{table}
\begin{table}[ht]
    \centering
    \caption{hyperparameter of Neat Resin, Random Forest, baseline}
    \label{tab:label_of_D3-1}
    \begin{tabular}{llllllll}
\toprule
$N_{\mathrm{train}}$ & 10 & 40 & 70 & 100 & 130 & 160 & 190 \\
\midrule
n\_estimators & 112 & 51 & 56 & 112 & 86 & 87 & 189 \\
max\_depth & 4 & 4 & 8 & 8 & 8 & 8 & 7 \\
min\_samples\_split & 2 & 8 & 7 & 3 & 3 & 2 & 2 \\
min\_samples\_leaf & 2 & 3 & 1 & 2 & 2 & 1 & 1 \\
\bottomrule
\end{tabular}
\end{table}
\begin{table}[ht]
    \centering
    \caption{hyperparameter of Neat Resin, Random Forest, Option1}
    \label{tab:label_of_D3-2}
    \begin{tabular}{llllllll}
\toprule
$N_{\mathrm{train}}$ & 10 & 40 & 70 & 100 & 130 & 160 & 190 \\
\midrule
n\_estimators & 178 & 136 & 200 & 97 & 182 & 170 & 165 \\
max\_depth & 3 & 7 & 5 & 7 & 5 & 5 & 7 \\
min\_samples\_split & 4 & 6 & 6 & 6 & 5 & 8 & 6 \\
min\_samples\_leaf & 2 & 2 & 2 & 3 & 3 & 1 & 3 \\
\bottomrule
\end{tabular}
\end{table}
\begin{table}[ht]
    \centering
    \caption{hyperparameter of Neat Resin, Random Forest, Option2}
    \label{tab:label_of_D3-3}
    \begin{tabular}{llllllll}
\toprule
$N_{\mathrm{train}}$ & 10 & 40 & 70 & 100 & 130 & 160 & 190 \\
\midrule
n\_estimators & 200 & 166 & 161 & 52 & 142 & 102 & 159 \\
max\_depth & 5 & 6 & 8 & 8 & 6 & 8 & 6 \\
min\_samples\_split & 2 & 3 & 2 & 2 & 2 & 2 & 6 \\
min\_samples\_leaf & 1 & 1 & 1 & 1 & 2 & 2 & 2 \\
\bottomrule
\end{tabular}
\end{table}
\begin{table}[ht]
    \centering
    \caption{hyperparameter of Neat Resin, Random Forest, Option3}
    \label{tab:label_of_D3-4}
    \begin{tabular}{llllllll}
\toprule
$N_{\mathrm{train}}$ & 10 & 40 & 70 & 100 & 130 & 160 & 190 \\
\midrule
n\_estimators & 63 & 50 & 59 & 196 & 88 & 123 & 70 \\
max\_depth & 2 & 4 & 6 & 8 & 6 & 7 & 8 \\
min\_samples\_split & 2 & 4 & 5 & 3 & 2 & 3 & 3 \\
min\_samples\_leaf & 1 & 1 & 3 & 2 & 2 & 3 & 1 \\
\bottomrule
\end{tabular}
\end{table}
\begin{table}[ht]
    \centering
    \caption{hyperparameter of Neat Resin, Random Forest, Option PolyCL}
    \label{tab:label_of_D3-5}
    \begin{tabular}{llllllll}
\toprule
$N_{\mathrm{train}}$ & 10 & 40 & 70 & 100 & 130 & 160 & 190 \\
\midrule
n\_estimators & 135 & 130 & 175 & 113 & 191 & 177 & 155 \\
max\_depth & 7 & 3 & 5 & 7 & 7 & 7 & 8 \\
min\_samples\_split & 2 & 6 & 3 & 4 & 2 & 4 & 6 \\
min\_samples\_leaf & 1 & 4 & 1 & 3 & 2 & 1 & 2 \\
\bottomrule
\end{tabular}
\end{table}
\begin{table}[ht]
    \centering
    \caption{hyperparameter of compound, Random Forest, baseline}
    \label{tab:label_of_D4-1}
    \begin{tabular}{llllllll}
\toprule
$N_{\mathrm{train}}$ & 10 & 40 & 70 & 100 & 130 & 160 & 190 \\
\midrule
n\_estimators & 52 & 184 & 54 & 128 & 179 & 197 & 105 \\
max\_depth & 2 & 3 & 5 & 7 & 6 & 6 & 8 \\
min\_samples\_split & 3 & 5 & 5 & 5 & 2 & 2 & 3 \\
min\_samples\_leaf & 1 & 1 & 1 & 2 & 2 & 1 & 1 \\
\bottomrule
\end{tabular}
\end{table}
\begin{table}[ht]
    \centering
    \caption{hyperparameter of compound, Random Forest, all features}
    \label{tab:label_of_D4-2}
    \begin{tabular}{llllllll}
\toprule
$N_{\mathrm{train}}$ & 10 & 40 & 70 & 100 & 130 & 160 & 190 \\
\midrule
n\_estimators & 168 & 189 & 50 & 143 & 191 & 77 & 136 \\
max\_depth & 7 & 2 & 4 & 4 & 6 & 5 & 7 \\
min\_samples\_split & 3 & 3 & 3 & 2 & 3 & 2 & 2 \\
min\_samples\_leaf & 2 & 2 & 2 & 2 & 2 & 2 & 1 \\
\bottomrule
\end{tabular}
\end{table}
\begin{table}[ht]
    \centering
    \caption{hyperparameter of compound, Random Forest, FS1}
    \label{tab:label_of_D4-3}
    \begin{tabular}{llllllll}
\toprule
$N_{\mathrm{train}}$ & 10 & 40 & 70 & 100 & 130 & 160 & 190 \\
\midrule
n\_estimators & 59 & 60 & 146 & 77 & 153 & 174 & 174 \\
max\_depth & 7 & 3 & 5 & 4 & 6 & 6 & 8 \\
min\_samples\_split & 4 & 4 & 6 & 3 & 5 & 4 & 2 \\
min\_samples\_leaf & 1 & 1 & 2 & 1 & 1 & 1 & 1 \\
\bottomrule
\end{tabular}
\end{table}
\begin{table}[ht]
    \centering
    \caption{hyperparameter of compound, Random Forest, FS2}
    \label{tab:label_of_D4-4}
    \begin{tabular}{llllllll}
\toprule
$N_{\mathrm{train}}$ & 10 & 40 & 70 & 100 & 130 & 160 & 190 \\
\midrule
n\_estimators & 83 & 53 & 145 & 84 & 183 & 83 & 131 \\
max\_depth & 4 & 2 & 8 & 5 & 8 & 6 & 8 \\
min\_samples\_split & 7 & 4 & 8 & 2 & 8 & 5 & 4 \\
min\_samples\_leaf & 1 & 3 & 2 & 1 & 2 & 1 & 1 \\
\bottomrule
\end{tabular}
\end{table}
\begin{table}[ht]
    \centering
    \caption{hyperparameter of compound, Random Forest, Option PolyCL}
    \label{tab:label_of_D4-5}
    \begin{tabular}{llllllll}
\toprule
$N_{\mathrm{train}}$ & 10 & 40 & 70 & 100 & 130 & 160 & 190 \\
\midrule
n\_estimators & 190 & 182 & 55 & 105 & 158 & 193 & 184 \\
max\_depth & 8 & 2 & 7 & 2 & 6 & 7 & 7 \\
min\_samples\_split & 2 & 3 & 5 & 6 & 5 & 8 & 4 \\
min\_samples\_leaf & 1 & 3 & 1 & 1 & 2 & 1 & 1 \\
\bottomrule
\end{tabular}
\end{table}

\clearpage
\section{Comparison between with/without hyperparameter tuning}\label{sec:appendix_e}
\begin{figure}[h]
 \begin{center}
 \begin{tabular}{c}
  \includegraphics[clip, width=1.0\columnwidth]
  {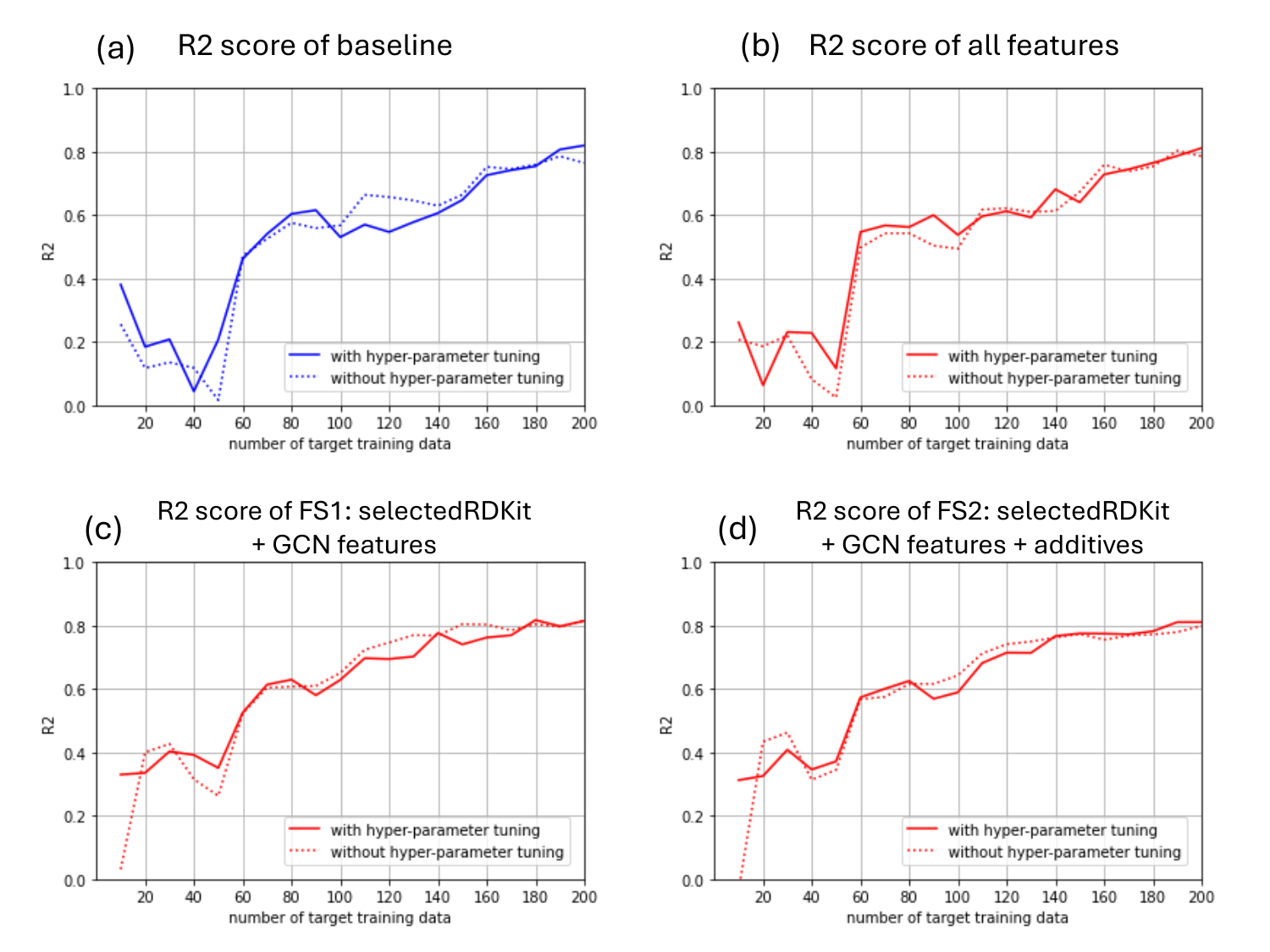}
 \end{tabular}
 \caption{Random Forest $R^2$ score of transferred to target neat resin datasets.
 left top: (a) baseline, right top: (b) All features, left down: (c) FS1: selected RDKit+GCN features, right down: (d) FS2: selected RDKit+GCN features+additives 
 \label{fig/mlst_E1_revised}
 }
 \end{center}
\end{figure}

\begin{figure}[h]
 \begin{center}
 \begin{tabular}{c}
  \includegraphics[clip, width=1.0\columnwidth]
  {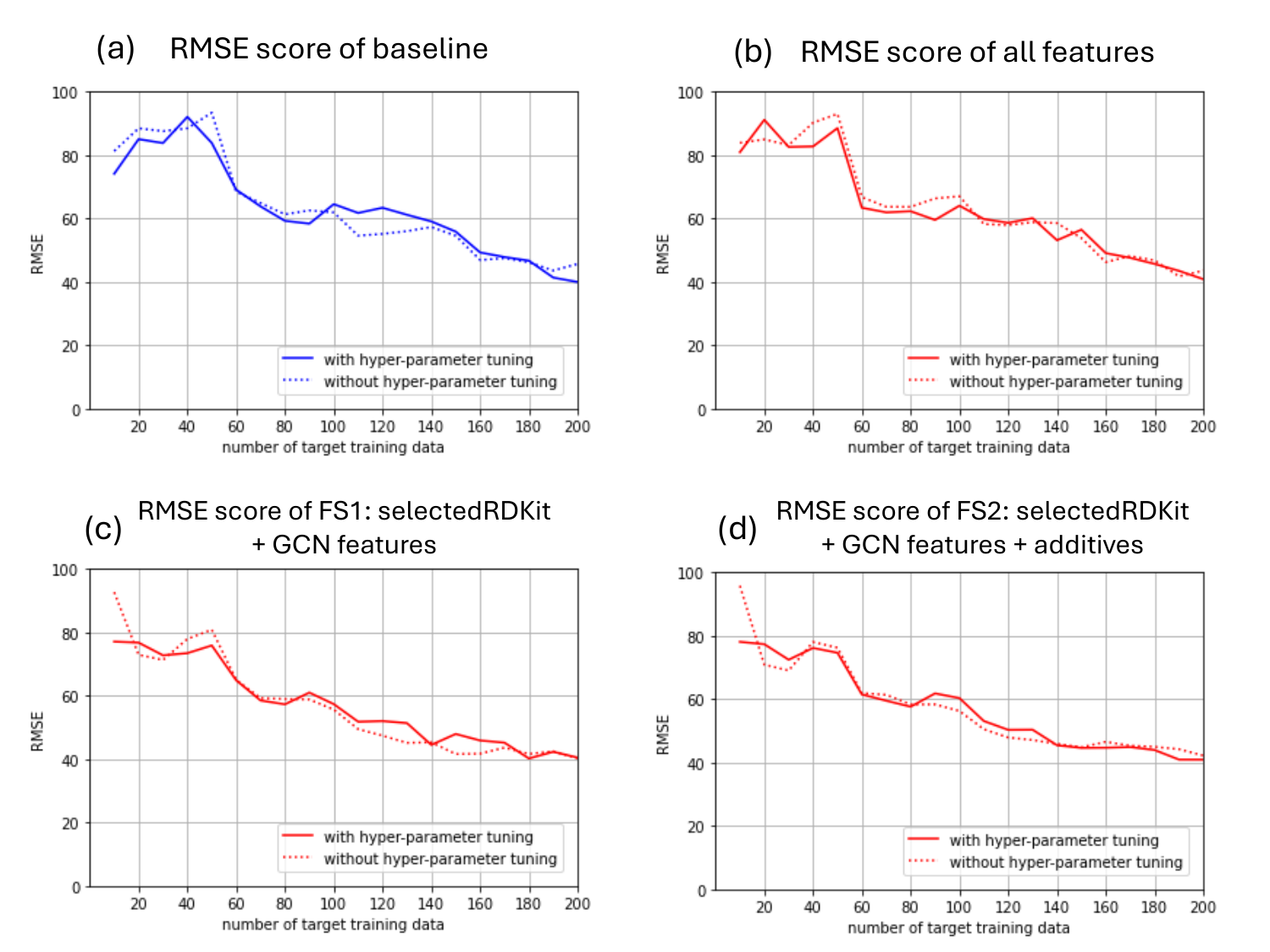}
 \end{tabular}
 \caption{Random Forest $RMSE$ score of transferred to target neat resin datasets.
left top: (a) baseline, right top: (b) All features, left down: (c) FS1: selected RDKit+GCN features, right down: (d) FS2: selected RDKit+GCN features+additives
 \label{fig/mlst_E2_revised}
 }
 \end{center}
\end{figure}

Following figures illustrate the comparison between the metrics score with hyperparameter tuning and the ones without hyperparameter tuning. ~\ref{fig/mlst_E1_revised} shows the $R^2$ score comparison of compound datasets. ~\ref{fig/mlst_E2_revised} shows the $RMSE$ score comparison of compound datasets. 


\end{document}